\documentclass{article} 
\usepackage{iclr2027_conference,times}

\usepackage{url,booktabs,amsmath,amssymb,amsfonts,amsthm,mathtools}
\usepackage{microtype,graphicx,flafter,placeins,xspace,xcolor,enumitem}
\usepackage{tabularx,array,multirow,seqsplit,wrapfig}
\usepackage{hyperref,cleveref}
\usepackage{pifont}
\newcommand{\rev}[1]{\textcolor{black}{#1}}

\title{When Rubrics Do Not Decide:\\Silent Value Completion in LLM Evaluation}
\author{Anonymous authors}
\date{}
\newcommand{\W}{\mathcal{W}}
\newcommand{\Spec}{\mathcal{S}}
\newcommand{\Cset}{\mathcal{C}}
\newcommand{\Fset}{\mathcal{F}}
\newcommand{\Wp}{\mathcal{W}_{\mathrm{p}}}
\newcommand{\Ws}{\mathcal{W}_{\mathrm{safe}}}
\newcommand{\R}{\mathbb{R}}
\DeclareMathOperator{\sign}{sign}
\newtheorem{proposition}{Proposition}
\Crefname{proposition}{Proposition}{Propositions}
\Crefname{appendix}{Appendix}{Appendices}

\title{When Rubrics Do Not Decide: \\ Silent Value Completion in LLM Evaluation}

\author{%
\makebox[\dimexpr\textwidth-2\tabcolsep\relax][c]{%
\begin{tabular}{@{}c@{}}
{\normalfont\normalsize
Zhuangzhuang Pan\textsuperscript{1}
\qquad
Yan Xia\textsuperscript{2,1}
\qquad
Kaiwen Lei\textsuperscript{1}
\qquad
Chee Seng Chan\textsuperscript{1,3*}}
\\[0.8em]
\resizebox{0.96\textwidth}{!}{%
\normalfont\small
\textsuperscript{1}Universiti Malaya, Malaysia
\quad
\textsuperscript{2}Suzhou University of Technology, China
\quad
\textsuperscript{3}VinUniversity, Vietnam}
\\[0.5em]
{%
\normalfont\small
\{23072126;23078403;24208178\}@siswa.um.edu.my} \\
{\normalfont\small
\textsuperscript{*}Corresponding author: cs.chan@um.edu.my}
\end{tabular}%
}%
}

\iclrfinalcopy

\begin{document}
\maketitle

\lhead{Preprint}

\begin{abstract}
\rev{Multi-criterion LLM judges often evaluate responses without specifying how conflicting criteria resolve, leaving the winner undetermined even under exact criterion scores.} \rev{Issuing a strict verdict without declaring the required trade-off premise constitutes \emph{silent value completion}.} \rev{We introduce \emph{TradeoffBench} to isolate this failure by separating rule execution from response scoring.} \rev{On authoritative criterion vectors, six open-weight non-reasoning judges pass all dominance and equality controls yet pick unsupported winners on 9\%--81\% of 1,345 conflicts.} \rev{All 577 order-stable choices follow an explicitly banned majority rule.} \rev{Reasoning-enabled configurations drop anonymous-vector violations to 2 in 9,720 judgments under an explicit rule, yet cannot bridge omitted trade-offs or divergent component scoring.} \rev{On raw TradeoffBench text with unspecified trade-offs, GPT-6.1-sol omits its completion premise in 29.8\% of 400 verified conflicts.} \rev{Even under an explicit rule on four-rater-consensus MultiPref conflicts, four reasoning judges still issue strict verdicts on 16/45 (GPT-6.1-sol) up to 35/46 and 25/29 (open-weight judges) of valid presentations.} \rev{We propose \emph{identified-first evaluation}, decoupling \emph{specification support}, \emph{explicit completion}, and \emph{target agreement}.} \rev{In MultiPref human-score audits, peer-mean substitution raises directional disagreement from 0.6\% to 28.8\% on 2,707 shared targets.} \rev{A single-peer ablation confirms a 29.38-point gap on 2,175 targets, ruling out averaging as the cause.} \rev{In HUMAINE, declared partial trade-offs add 2.5 percentage points of coverage, yet newly decided pairs incur 3.9\% disagreement versus 0.9\% for Pareto.} \rev{Reliable LLM judging therefore requires auditing specification support, declared trade-offs, and human target agreement separately.}
\end{abstract}

\section{Introduction}
\label{sec:intro}

\begin{figure}[t]
\centering
\includegraphics[width=0.7\linewidth]{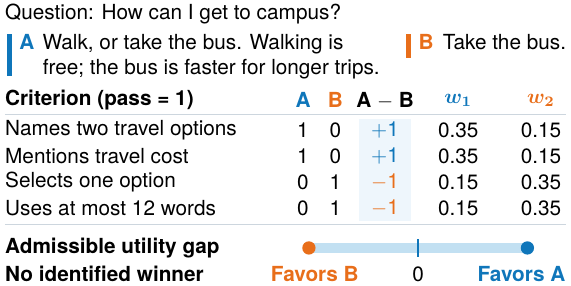}
\caption{\rev{Travel-planning example of rubric underdetermination, with binary criterion scores of pass (1) or fail (0).} \rev{Every weighting between $\boldsymbol{w}_1$ and $\boldsymbol{w}_2$ satisfies the rubric.} \rev{Because the utility-difference interval spans both positive and negative values, the rubric determines no winner.}}
\label{fig:conflict}
\vspace{-15pt}
\end{figure}

\rev{An LLM judge can agree with a human evaluator even when the grading rubric leaves the winner undetermined.} \rev{This ambiguity arises when a rubric defines individual criteria without specifying how conflicts between them resolve.} \rev{Figure~\ref{fig:conflict} illustrates this failure on a travel-planning prompt, where each response receives binary pass or fail scores on every criterion.} \rev{Response A provides two travel options with cost details, whereas Response B offers one concise recommendation.} \rev{Under weighting $w_1$, Response A wins with a score of $0.70$ versus $0.30$; under weighting $w_2$, Response B wins by the same margin.} \rev{Because both weightings satisfy the rubric, neither winner is justified without a declared trade-off rule.} \rev{We define \emph{silent value completion} as issuing a strict verdict without declaring the required additional premise.} \rev{This term describes an unsupported decision rather than a judge's internal reasoning or intent.}

\rev{Distinguishing human agreement from rubric support matters because existing evaluations rarely test whether verdicts follow from the stated rubric.} \rev{Rubric-based methods formalize evaluation criteria and generate structured checklist feedback \citep{ye2024flask,kim2024prometheus,liu2024alignbench,kim2024prometheus2,viswanathan2025checklists,roy2026premise}.} \rev{Benchmark studies audit human agreement, position bias, style sensitivity, and rubric artifacts \citep{zheng2023judging,wang2024fair,feuer2025style,bagaria2026rubricartifacts}.} \rev{Yet a judge can deliver high agreement while silently trading off criteria such as safety against concision without an audit trail.}

\rev{To isolate this behavior, we introduce \emph{TradeoffBench}, which separates rule execution from response scoring.} \rev{Its controlled vector audit scores five binary criteria exactly under an explicit strict Pareto rule.} \rev{Under this rule, dominance requires a strict winner, identical profiles require an equality label, and conflicting criteria require deferral.} Additional weights, priorities, and majority voting are prohibited.

\rev{In non-reasoning configurations, six open-weight judges handle all dominance and equality controls yet pick unsupported winners on 9\%--81\% of 1,345 original-order conflicts.} \rev{All 577 order-stable unsupported picks follow the banned criterion majority.} \rev{On an exhaustive $3^5$ profile grid with higher token budgets, reasoning-enabled configurations cut vector violations to 2 in 9,720 judgments (\Cref{sec:configuration}).} \rev{On raw response text under unspecified trade-offs, however, GPT-6.1-sol still leaves completion premises unstated in 29.8\% of verified-conflict presentations (\Cref{sec:raw-open}).}

\rev{These diagnostics reveal a fundamental boundary between following a decision rule and having sufficient information to decide.} \rev{Better reasoning improves rule execution, yet cannot resolve a trade-off the rubric omits.} \rev{Furthermore, even a mathematically supported verdict can disagree with the intended human evaluator when input component scores differ from that rater's assessments.}

\rev{We therefore propose \emph{identified-first evaluation}, organizing evaluation around three distinct requirements.} \rev{\emph{Specification support} identifies strict verdicts entailed by every admissible weighting, following necessary preference in robust ordinal regression \citep{greco2008ordinal}.} \rev{\emph{Explicit completion} requires evaluators to declare any trade-off premise used to resolve undecided pairs.} \rev{\emph{Target agreement} audits how often the resulting decisions match the intended human rater.} \rev{Decoupling these three stages makes it possible to track what the rubric justifies, what added premises decide, and how often either aligns with people.}

\rev{We test target agreement and explicit completion directly on human criterion scores without an LLM judge.} \rev{In MultiPref, replacing raters' own scores with peer means raises directional disagreement from 0.6\% to 28.8\% on 2,707 shared comparisons.} \rev{A single-peer ablation confirms a 29.38-point increase on 2,175 strict targets.} \rev{In HUMAINE, declared partial trade-offs raise coverage by 2.5 percentage points over strict Pareto dominance.} \rev{However, newly decided pairs incur higher disagreement than Pareto pairs (3.9\% versus 0.9\%).} \rev{Together, these results show that correct rule execution does not ensure agreement with the intended human evaluator.}

\rev{This paper makes three main contributions.}
\begin{enumerate}[leftmargin=*,label=\arabic*.,itemsep=0pt,topsep=0pt]
\item \rev{We characterize silent value completion and introduce \emph{TradeoffBench} to separate rule execution from response scoring.} \rev{Controlled and raw-text audits show that non-reasoning judges invent majority rules on conflicts, while a frontier reasoning judge still omits trade-off premises on response text under unspecified rubrics.}

\item \rev{We formulate \emph{identified-first evaluation} and demonstrate on MultiPref that component-source substitution degrades target agreement even under exact rule execution.} \rev{This finding establishes why target agreement requires an audit separate from specification support.}

\item \rev{We quantify the coverage and risk of explicit completion on HUMAINE using declared partial trade-offs.} \rev{Our qualification audit isolates higher disagreement on newly decided pairs while using the non-identified frontier to prioritize human review.}
\end{enumerate}

\section{Identified-First Evaluation: Three Boundaries}
\label{sec:method}

\rev{\emph{Identified-first evaluation} prevents silent value completion while keeping human agreement as a separate evaluation objective.} \rev{First, specification support identifies which strict verdicts the declared rubric justifies (\Cref{sec:specification}).} \rev{Second, explicit completion records any additional trade-off premise required to resolve undecided pairs (\Cref{sec:core-frontier}).} \rev{Third, target agreement measures how often the resulting decisions match the intended human evaluator (\Cref{sec:calibration}).}

\subsection{Specification Support: What Does the Rubric Justify?}
\label{sec:specification}

\rev{We represent the declared specification as $\Spec=(\mathcal D,\phi,\W)$.} \rev{Here $\mathcal D$ lists the $d$ evaluation criteria, and $\phi$ maps candidate pairs to criterion differences $x\in\R^d$.} \rev{Positive $x_k$ favors Response A, negative $x_k$ favors Response B, and zero denotes equality on criterion $k$.} \rev{The admissible trade-off set collects all allowed weightings}
\begin{equation}
\W=\{w\in\R_+^d:\mathbf1^\top w=1,\;Aw\leq b\},
\label{eq:weight-set}
\end{equation}
\rev{which we assume to be nonempty.} \rev{While $\phi$ fixes each criterion's direction and scale, $\W$ specifies the permissible compensations across criteria.}

\rev{A strict verdict has \emph{specification support} only when every admissible weighting favors the same candidate.} \rev{We compute the interval of possible utility differences across $\W$ as}
\begin{equation}
[\underline u_\W(x),\overline u_\W(x)]
=
[\min_{w\in\W}w^\top x,\;
 \max_{w\in\W}w^\top x].
\label{eq:robust-interval}
\end{equation}
\rev{In our travel-planning example (\Cref{fig:conflict}), Response A scores $0.70$ versus $0.30$ under $\boldsymbol{w}_1$, while Response B wins by the same margin under $\boldsymbol{w}_2$.} \rev{Because the interval spans both positive and negative values, the rubric justifies neither winner.}

\rev{Applying necessary preference from robust ordinal regression \citep{greco2008ordinal}, we define the identified-first rule as}
\begin{equation}
\begin{gathered}
r_\Spec(x)=
\begin{cases}
+1,&\underline u_\W(x)>0,\\
-1,&\overline u_\W(x)<0,\\
0,&\text{otherwise},
\end{cases}\\
\Cset_\Spec=\{x:r_\Spec(x)\neq0\},\qquad
\Fset_\Spec=\{x:r_\Spec(x)=0\}.
\end{gathered}
\label{eq:core}
\end{equation}
\rev{The \emph{identified core} $\Cset_\Spec$ contains comparisons whose strict winner holds across every weighting in $\W$.} \rev{All remaining cases fall into the \emph{non-identified set} $\Fset_\Spec$, where the rule withholds a strict decision.}

\rev{Cases inside $\Fset_\Spec$ differ conceptually.} \rev{An interval containing both signs allows opposing winners, whereas an interval touching zero permits a weak direction without strict dominance.} \rev{An interval equal to $[0,0]$ establishes exact model equality, yet still does not mean the human evaluator is indifferent.} \rev{None of these three outcomes supports a strict release under \Cref{eq:core}.}

\rev{When $\W$ is polyhedral, two linear programs compute $[\underline u_\W(x),\overline u_\W(x)]$.} \rev{Arbitrarily small or zero weights in $\R_+^d$ can suppress strict dominance, so Pareto-consistent specifications either impose strictly positive lower bounds or declare the strict Pareto rule directly.} \rev{Two structural properties formalize how $r_\Spec$ behaves across weight sets, with proofs in \Cref{app:proofs}.}

\begin{proposition}[\rev{Nesting and compatibility}]
\label{prop:nesting}
\rev{Hold the criteria and coding fixed. If $\W_1\subseteq\W_2$, then $\Cset_{\Spec_2}\subseteq\Cset_{\Spec_1}$ and strict directions agree. If every $w\in\W$ is strictly positive, the core contains every strict Pareto direction. If $w_0\in\W$, each core direction agrees with $\sign(w_0^\top x)$.}
\end{proposition}

\begin{proposition}[\rev{Pointwise maximality within the soundness class}]
\label{prop:maximality}
\rev{Call $h$ specification-sound when $h(x)=s\in\{-1,+1\}$ implies $s\,w^\top x>0$ for every $w\in\W$. Every strict release of such an $h$ is released in the same direction by $r_\Spec$.}
\end{proposition}

\rev{Stakeholders determine the criteria, scales, and admissible trade-offs.} \rev{They can supply explicit weight bounds or reference comparisons to constrain $\W$ \citep{greco2008ordinal}.} \rev{We evaluate reference-comparison elicitation on individual MultiPref histories in \Cref{sec:elicitation} (\Cref{app:elicitation}).}

\subsection{Explicit Completion: How Can Undecided Comparisons Be Resolved?}
\label{sec:core-frontier}

\rev{Specification support identifies what the rubric justifies, yet it does not ensure a winner on every pair.} \rev{When an application requires a strict choice, the evaluator may supply an \emph{explicit completion}.} \rev{This step adds a declared trade-off premise that restricts $\W$ to a nonempty subset $\W_p\subseteq\W$.}

\begin{figure}[t]
\centering
\includegraphics[width=\textwidth]{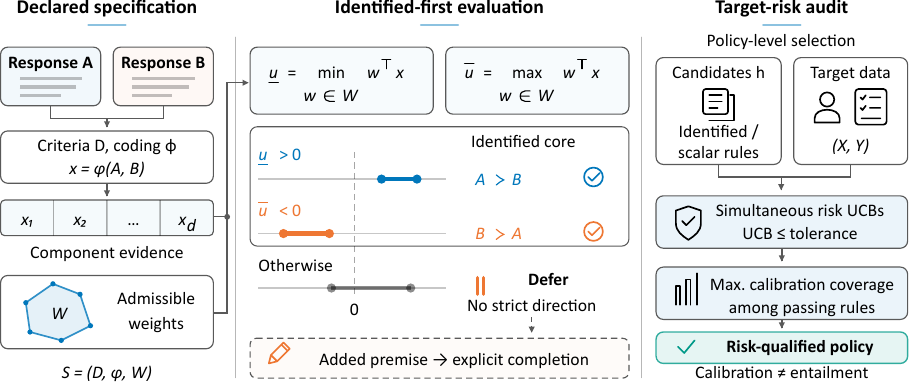}
\caption{Identified-first evaluation separates three requirements. Specification support identifies verdicts justified by the declared rubric. Explicit completion records additional trade-off premises. Target-risk auditing evaluates the resulting decisions against the intended rater.}
\label{fig:overview}
\vspace{-15pt}
\end{figure}

\rev{Restricting $\W$ can resolve mixed-sign conflicts through the added premise rather than the original rubric.} \rev{If the initial utility interval equals $[0,0]$, however, narrowing $\W$ alone yields no strict direction.} \rev{Deciding an equal-utility pair requires an assumption outside the additive specification.} \rev{\Cref{fig:overview} traces how specification support, explicit completion, and target-risk auditing fit together.}

\rev{Picking the single equal-weight vector $w_0=\mathbf1/d$, when admissible, illustrates explicit completion.} \rev{Its utility difference equals $w_0^\top x=d^{-1}\sum_k x_k$, selecting the criterion majority on common-scale ternary scores.} \rev{That choice reflects a declared weighting rather than an implicit mandate of the original rubric.}

\rev{By contrast, TradeoffBench and MultiPref directly use the strict Pareto rule}
\begin{equation}
r_{\mathrm P}(x)=
\begin{cases}
+1,&x\geq0,\ x\neq0,\\
-1,&x\leq0,\ x\neq0,\\
0,&\text{otherwise}.
\end{cases}
\label{eq:pareto-rule}
\end{equation}
\rev{Under \Cref{eq:pareto-rule}, a winner must improve on at least one criterion without worsening any other, forcing deferral on every conflict.} \rev{Because arbitrary nonnegative weights in $\R_+^d$ could assign zero weight to every improved criterion, strict Pareto requires positive weight lower bounds when derived from $\W$.} \rev{HUMAINE's partial specifications impose such positive bounds explicitly to expand strict decisions beyond dominance.}

\rev{The operational boundary therefore lies between verdicts justified by the original specification and verdicts justified only after adding a premise.} \rev{Retaining that boundary lets evaluators track both coverage and risk separately.}

\subsection{Target Agreement: Does a Supported Verdict Match the Intended Rater?}
\label{sec:calibration}

\rev{A verdict can follow the declared specification yet disagree with the intended human evaluator.} \rev{Conversely, a verdict relying on explicit completion may match that evaluator.} \rev{Neither specification support nor accurate rule execution ensures target agreement.}

Let $Y\in\{-1,0,+1\}$ denote the intended rater's overall judgment, where zero represents a target tie. \rev{For a decision policy $h$ whose zero output denotes deferral, we measure strict-direction risk and all-record coverage through}
\begin{equation}
R_{\rm dir}(h)=
\Pr\{h(X)\neq Y\mid h(X)\neq0,\;Y\neq0\},
\qquad
C(h)=\Pr\{h(X)\neq0\}.
\label{eq:strict-risk}
\end{equation}

\rev{Strict-direction risk captures disagreement among released decisions that have strict human targets.} \rev{All-record coverage additionally counts comparisons with target ties.} \rev{We also track tie-aware risk, $R_{\rm tie}(h)=\Pr\{h(X)\neq Y\mid h(X)\neq0\}$, which penalizes any strict verdict on a tied target as an error.} \rev{Both conditional risks remain undefined when their conditioning events carry zero probability.}

\rev{These definitions motivate two complementary empirical audits.} \rev{First, MultiPref isolates how replacing an evaluator's criterion scores affects disagreement while holding the intended rater and decision rule fixed (\Cref{sec:result-alignment}).} \rev{Second, HUMAINE measures how declared partial trade-offs raise coverage and target risk across newly decided cases (\Cref{sec:result-humaine}).} \rev{Together, both studies test failure modes that rule compliance alone cannot resolve.}

\paragraph{Execution with either inference mode.}
\label{par:mode-independent-execution}
\rev{Given structured criterion scores, the evaluator computes the supported verdict under the declared rule and logs any added premise before recomputation.} \rev{Strict Pareto executes in $O(d)$ operations, whereas general polyhedral specifications solve two linear programs.} \rev{This deterministic step audits specification support rather than component validity or target agreement.} \rev{For already compliant judges, it records explicit support without changing any verdict (\Cref{app:reasoning-integration}).}

\section{TradeoffBench: Strict Verdicts Beyond the Declared Rule}
\label{sec:experiments}
\label{sec:result-audit}

\rev{Before evaluating how closely an LLM judge matches human preferences, we test whether the judge respects the rubric's own logical boundaries.} \rev{When a rubric defines criteria without specifying trade-offs, a mathematically sound evaluator must withhold a strict verdict.} \rev{We introduce \emph{TradeoffBench} to separate rule execution from text-comprehension errors.}

\subsection{Verified comparisons and a controlled decision rule}
\rev{TradeoffBench generates addition, comparison, parity, arithmetic-sequence, and unit-conversion tasks across five binary criteria.} \rev{Those criteria cover correctness, completeness, format, concision, and lexical compliance (using a required word).} \rev{Four stochastic response policies produce six candidate pairs per latent task, and a deterministic validator ensures each rendered text matches its sampled profile.} \rev{This setup extends verifiable instruction constraints \citep{zeng2024llmbar,jiang2024followbench,qin2024infobench} to an aggregation audit with known ground-truth relations.}

\rev{Consider comparing 66 against 11, where both responses answer correctly and pass the format and lexical checks.} \rev{Response A alone passes completeness, whereas Response B alone passes concision.} \rev{In the vector \texttt{[SAME, A\_ONLY, SAME, B\_ONLY, SAME]}, \texttt{SAME} indicates equality while \texttt{A\_ONLY} and \texttt{B\_ONLY} identify the winning candidate.} \rev{Because both candidates hold advantages, the rule requires \textsc{underdetermined} (\Cref{app:tradeoff-example}).}

\rev{The matched audit spans 4,200 pairs from 700 latent tasks, comprising 2,566 dominance controls, 289 equalities, and 1,345 mixed-sign conflicts.} \rev{Every controlled prompt supplies an exhaustive rule forbidding inferred weights, priorities, severity overrides, and majority voting.} \rev{Dominance requires A or B, all-equal profiles require equality, and conflicts require deferral.}

\rev{Our $2\times2$ factorial design varies both evidence modality and label separation across fixed tasks and criterion orders.} \rev{\emph{Raw evidence} supplies the question, rubric, and response texts, whereas \emph{Vector evidence} supplies only authoritative \texttt{A\_ONLY}, \texttt{B\_ONLY}, and \texttt{SAME} tokens.} \rev{\emph{Combined labels} merge equality and conflict under \textsc{tie}, whereas \emph{Disjoint labels} separate \textsc{equal} from \textsc{underdetermined}.} \rev{All six checkpoints run under greedy decoding with official chat templates, with reasoning disabled for Qwen3-8B and EXAONE-4.0-32B (\Cref{app:tradeoff}).}

\subsection{Exact vectors leave unsupported strict choices}
\label{sec:endpoint}

\begin{wraptable}[11]{R}{0.50\textwidth}
\vspace{-10pt}
\setlength{\abovecaptionskip}{0pt}
\caption{\rev{Unsupported strict-verdict rates on 1,345 mixed-sign pairs at Vector/disjoint; 95\% confidence intervals (CIs) use a latent-task bootstrap.}}
\label{tab:endpoint-main}
\centering\footnotesize
\setlength{\tabcolsep}{2pt}
\begin{tabular*}{\linewidth}{@{\extracolsep{\fill}}lrr@{}}
\toprule
Model & $n$ & Rate (\%) [95\% CI]\\
\midrule
Qwen3-8B & 345 & 25.65 [23.27, 28.03]\\
Mistral-Small-24B & 119 & 8.85 [7.27, 10.48]\\
Phi-4-14B & 301 & 22.38 [20.10, 24.69]\\
OLMo-Hybrid-7B & 237 & 17.62 [15.49, 19.82]\\
GLM-4-9B & 1,091 & 81.12 [79.00, 83.20]\\
EXAONE-4.0-32B & 140 & 10.41 [8.63, 12.24]\\
\bottomrule
\end{tabular*}
\end{wraptable}

\rev{At Vector/disjoint, all six models pick unsupported winners on 8.85\% to 81.12\% of the 1,345 conflicts (\Cref{tab:endpoint-main}).} \rev{Rates range from Mistral-Small-24B (119 pairs) to GLM-4-9B (1,091 pairs).} \rev{Even so, every checkpoint scores 100.0\% on all 2,566 dominance and 289 equality controls, with all 4,200 outputs parseable per model.} \rev{Removing response text and separating conflict labels from equality therefore does not eliminate rule violations.}

\rev{For secondary factorial and ordinary-interface diagnostics, we partition the 1,345 conflicts using an external ordered weight set that remains undisclosed to the models.} \rev{That external set leaves 849 benchmark-frontier pairs unresolved while ordering the other 496 conflicts (\Cref{app:controlled-audit}).} \rev{With disjoint labels fixed, switching from raw text to exact vectors reduces strict choices in every model.} \rev{However, raw conditions achieve only 44.6\% to 75.3\% dominance accuracy and therefore confound silent completion with broader rule-execution failures.} \rev{Our primary endpoint retains all 1,345 conflicts to preserve benchmark frequencies, with uncertainty resampled over latent tasks (\Cref{sec:artifact}).}

\subsection{Candidate-consistent strict choices follow the majority direction}
\label{sec:endpoint-swap}
\rev{To separate preference-like completions from label position bias, we test all 4,200 Vector/disjoint pairs in both A/B orientations.} \rev{We exchange only \texttt{A\_ONLY} and \texttt{B\_ONLY}, keeping criterion order, instructions, checkpoint, and greedy decoding fixed.}

\rev{We call a pair \emph{candidate-consistent} when both orientations emit strict verdicts selecting the same underlying response after label mapping.} \rev{Pairs emitting A/A or B/B select opposite responses across orders and are \emph{position-locked}, while remaining pairs emit a strict verdict once or never.} \rev{Every dominance and equality control remains 100.0\% accurate in both orientations.} \rev{Across the five models with nonzero stability, candidate-consistent rates span 4.61\%--13.09\% of the 1,345 conflicts, from OLMo-Hybrid-7B (62 pairs) to GLM-4-9B (176 pairs, \Cref{tab:swap-main}).}

\begin{table}[t]
\vspace{-10pt}
\caption{A/B-exchange diagnostic at Vector/disjoint. Counts partition all 1,345 mixed-sign pairs into candidate-consistent, position-locked (A/A or B/B), one-strict, and neither-strict outcomes. Consistency rates use all pairs; brackets give pointwise 95\% latent-task bootstrap CIs.}
\label{tab:swap-main}
\centering\footnotesize
\setlength{\tabcolsep}{4pt}
\begin{tabular*}{\textwidth}{@{\extracolsep{\fill}}lrrrrr@{}}
\toprule
& \multicolumn{2}{c}{Candidate-consistent} & \multicolumn{3}{c}{Other paired outcomes ($n$)}\\
\cmidrule(lr){2-3}\cmidrule(lr){4-6}
Model & $n$ & Rate (\%) [95\% CI] & Position-locked & One strict & Neither strict\\
\midrule
Qwen3-8B & 72 & 5.35 [4.02, 6.76] & 78 & 396 & 799\\
Mistral-Small-24B & 0 & 0.00 [--] & 0 & 248 & 1,097\\
Phi-4-14B & 165 & 12.27 [10.47, 14.13] & 0 & 272 & 908\\
OLMo-Hybrid-7B & 62 & 4.61 [3.44, 5.89] & 11 & 335 & 937\\
GLM-4-9B & 176 & 13.09 [11.28, 14.99] & 766 & 319 & 84\\
EXAONE-4.0-32B & 102 & 7.58 [6.10, 9.13] & 0 & 82 & 1,161\\
\bottomrule
\end{tabular*}
\vspace{-15pt}
\end{table}

\rev{Crucially, all 577 candidate-consistent model--pair choices favor the banned criterion majority, $\sign(\sum_k x_k)$.} \rev{Agreement holds at 100\% under both pair-frequency and distinct-profile weighting.} \rev{Because the same conflicts recur across models, these represent 577 model--pair events rather than independent observations, and the pattern reflects output behavior without identifying an internal weighting mechanism.}

\rev{Vote structure further separates stability from order sensitivity.} \rev{Among 827 strict-majority and 518 tied-vote pairs, zero tied pairs become candidate-consistent.} \rev{Mistral-Small-24B and EXAONE-4.0-32B defer on all 1,036 tied-vote presentations, whereas the other four models' occasional strict picks on tied votes always favor displayed A.} \rev{Even a zero consistency rate fails to prove position invariance.} \rev{Mistral-Small-24B picks strict majority winners 119 times in the original order and 129 times after swapping, occurring on disjoint pairs whenever that majority candidate appears as displayed A (\Cref{app:endpoint-swap}).}

\subsection{Near-zero rule violations in tested reasoning-enabled configurations}
\label{sec:configuration}
\rev{Testing reasoning-enabled configurations across all $3^5$ profiles and three anonymous encodings reduces rule violations to 2 in 9,720 mixed-sign orientation judgments (\Cref{app:lattice-audit}).} \rev{All 3,402 dominance and equality controls pass at 100\%, ruling out an always-defer trivial strategy.} \rev{This exhaustive grid removes benchmark-frequency weighting and evaluates Qwen3-8B, EXAONE-4.0-32B, and Qwen3.8-27B.}

\rev{By contrast, matched reasoning-disabled runs generate 376 candidate-consistent majority events across 74 canonical profiles, including on signed-numeric vectors, with zero on the 50 tied-vote profiles.} \rev{Six of the nine reasoning-disabled cells also fail dominance controls (332/372 to 366/372), confirming broader execution errors alongside silent completion.}

\rev{All within-model comparisons in \Cref{app:lattice-audit} hold sampling parameters, random seeds, and the 2,048-token cap fixed while switching official reasoning-mode settings.} \rev{Those comparisons isolate configuration behavior rather than an internal reasoning mechanism, and the higher token cap differs from the greedy primary audit in \Cref{sec:endpoint}.} \rev{Observed maximum completions rise from 16--17 tokens in reasoning-disabled mode to 293--1,511 tokens with reasoning enabled (\Cref{app:lattice-audit}).} \rev{Because correct deferral still leaves conflicts unresolved, Sections~\ref{sec:result-alignment} and~\ref{sec:result-humaine} remove the LLM judge and apply declared rules directly to human scores.}

\subsection{Raw responses and open-ended rubrics}
\label{sec:raw-open}
\rev{We next evaluate reasoning-enabled Qwen3-8B, Qwen3.8-27B, and the frontier judge GPT-6.1-sol directly on raw response text, requiring joint component assessment and aggregation.} \rev{Each judge evaluates 400 TradeoffBench pairs in both orders, comprising 200 verified conflicts and 100 each of dominance and equality controls.} \rev{Under explicit-rule prompting, strict conflict verdicts total 84/400 for Qwen3-8B, 31/400 for Qwen3.8-27B, and 0/400 for GPT-6.1-sol.} \rev{Control accuracy reaches 396/400, 400/400, and 400/400, respectively (\Cref{app:raw-open}).}

\begin{table}[!t]
\caption{\rev{GPT-6.1-sol on the same 800 raw TradeoffBench presentations per condition, comprising 400 conflicts and 400 controls.} \rev{Reported conflicts and unstated premises are conditioned on strict verdicts; explicit-rule outputs return only a verdict.}}
\label{tab:raw-open}
\centering\footnotesize
\begin{tabular*}{\textwidth}{@{\extracolsep{\fill}}lrrrr@{}}
\toprule
Trade-offs & \shortstack{Strict on\\verified conflicts} & \shortstack{Reported\\conflict} & \shortstack{Unstated\\premise} & \shortstack{Correct\\controls}\\
\midrule
Explicit rule & 0 / 400 & --- & --- & 400 / 400\\
Unspecified & 176 / 400 & 176 / 176 & 119 / 176 & 400 / 400\\
\bottomrule
\end{tabular*}
\vspace{-15pt}
\end{table}

\rev{When trade-offs remain unspecified on the same raw pairs, GPT-6.1-sol still issues 176/400 strict conflict verdicts (44.0\%; 95\% CI $[37.6,50.7]$, \Cref{tab:raw-open}).} \rev{Its own reported components retain all 400 conflicts, and 153 of the 176 strict verdicts follow complete matches to the verified vector.} \rev{Yet 119 of those 176 strict verdicts omit any completion premise, corresponding to 29.8\% of the 400 verified conflicts and 67.6\% of strict verdicts.} \rev{Fifty-two of those unstated-premise responses contain only components and a verdict, including 49 with exact verified vectors.} \rev{Although both conditions match presentations and inference parameters, the instruction and output format change together (\Cref{app:raw-open}).}

\rev{For open-ended judging, GPT-6.1-sol evaluates all 73 MultiPref pairs where four raters agree on the full helpfulness, truthfulness, and harmlessness sign vector.} \rev{Four conditions cross model-assessed versus fixed human-reference components against unspecified trade-offs versus the explicit rule, allowing explanations and deferral throughout (\Cref{app:realtext-diagnostic}).} \rev{Among 36 strict verdicts issued over reported conflicts under unspecified trade-offs, 29 state a premise, six leave it unstated, and one is ambiguous.} \rev{Two unstated-premise outputs match the full four-rater vector, whereas the explicit rule yields zero strict verdicts on any reported conflict.}

\rev{Even under the explicit strict Pareto rule, all four reasoning judges still emit strict verdicts on 16/45 (GPT-6.1-sol), 25/46 (Qwen3-8B), 35/46 (Qwen3.8-27B), and 25/29 (Qwen3.5-9B) valid presentations of the 23 four-rater-consensus conflict pairs (\Cref{tab:consensus-measurement}).} \rev{Every strict verdict follows dominance in the model's own reported components, with zero rule mismatches across all four judges.} \rev{Explicit instructions and strong reasoning therefore enforce internal rule compliance without preventing spurious winners when component assessment diverges from human raters.}

\begin{table}[!t]
\caption{Model-assessed components under the explicit strict Pareto rule on 23 four-rater-consensus MultiPref conflict pairs, each in both orders. Counts use valid outputs under each audit's primary parsing policy. Reported conflicts refer to the model's components, while rule mismatches compare the verdict against the rule applied to those components.}
\label{tab:consensus-measurement}
\centering\footnotesize
\setlength{\tabcolsep}{4pt}
\begin{tabular*}{\textwidth}{@{\extracolsep{\fill}}lrrrrr@{}}
\toprule
Judge & \shortstack{Valid /\\planned} & \shortstack{Strict\\verdicts} & \shortstack{Reported\\conflicts} & EQUAL & \shortstack{Rule\\mismatches}\\
\midrule
GPT-6.1-sol & 45 / 46 & 16 & 29 & 0 & 0\\
Qwen3.5-9B & 29 / 46 & 25 & 4 & 0 & 0\\
Qwen3.8-27B & 46 / 46 & 35 & 10 & 1 & 0\\
Qwen3-8B & 46 / 46 & 25 & 21 & 0 & 0\\
\bottomrule
\end{tabular*}
\vspace{-10pt}
\end{table}

\section{MultiPref: Identification Does Not Ensure Target Agreement}
\label{sec:result-alignment}

\rev{Section~\ref{sec:result-audit} showed that LLM judges often fail specification support by silently inventing trade-offs.} \rev{Removing the LLM isolates whether flawless rule execution alone aligns evaluation verdicts with human raters.} \rev{Using MultiPref \citep{miranda2024hybrid}, we test whether component-source substitution breaks target agreement even under an exact strict Pareto rule, complementing our LLM-assessed confirmation checks in \Cref{app:workflow}.}

\rev{In MultiPref, four evaluators judge each comparison on helpfulness, truthfulness, and harmlessness \citep{miranda2024hybrid}.} \rev{Component preferences take values in $\{-1,-0.5,0,0.5,1\}$, while each evaluator's overall label remains fixed.} \rev{We apply \Cref{eq:pareto-rule} identically across two conditions using either the target evaluator's own components or peer-mean components averaged coordinatewise from the other three raters, treating zero as coordinate equality.}

\rev{Applying a global-time split and excluding development prompts yields 4,847 evaluation decisions from 80 eligible evaluators, including 3,515 strict overall targets.} \rev{Earlier data sets evaluator eligibility without fitting any completion rule.} \rev{Because substitution alters which comparisons enter the core, our main paired contrast holds fixed the 2,707 strict targets released in both conditions (\Cref{app:multipref}).}

\begin{figure}[t]
\centering
\includegraphics[width=0.95\textwidth]{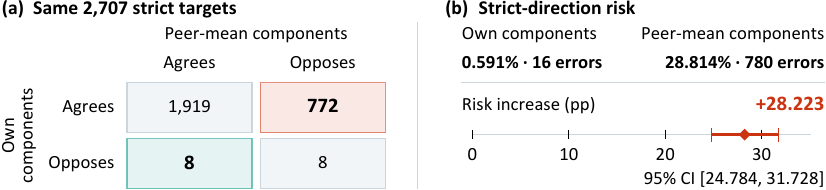}
\caption{Component substitution on the same strict targets, with target labels and the strict Pareto rule fixed. The diamond and whisker show the paired risk increase and 95\% evaluator--prompt crossed-bootstrap CI. Counts are in \Cref{tab:transport-all}.}
\label{fig:transport-main}
\vspace{-15pt}
\end{figure}

\rev{Replacing raters' own components with peer means raises strict-direction risk from 0.591\% to 28.814\% across those 2,707 shared targets, increasing errors from 16 to 780 (\Cref{fig:transport-main} and \Cref{tab:transport-all}).} \rev{Every decision still follows strict Pareto for its supplied inputs with the target fixed, yet disagreement jumps by 28.223 percentage points.} \rev{Rule support and target agreement therefore require separate verification.}

\rev{We estimate uncertainty using 20,000 crossed-bootstrap draws over evaluators and prompts while preserving within-record pairing.} \rev{Across all eligible records, own components yield lower residual error than peer means for 73 of 80 evaluators, equal error for seven, and higher error for none.} \rev{The earlier partition shows a consistent gap of 0.724\% versus 29.637\% (\Cref{app:multipref-consistency}).} \rev{To test whether peer averaging causes this gap, a single-peer ablation fixes 2,175 strict targets released across all three conditions (\Cref{app:single-peer}).} \rev{Single-peer substitution raises disagreement over raters' own components by 29.38 points (three-contrast-adjusted CI $[24.75,34.10]$), showing that the source gap persists without averaging.} \rev{Both analyses measure alignment with the intended rater rather than objective deployment accuracy.}

\section{\texorpdfstring{\rev{Explicit Trade-offs: HUMAINE Qualification and Weight Elicitation}}{Explicit Trade-offs: HUMAINE Qualification and Weight Elicitation}}
\label{sec:result-humaine}

\rev{When applications require strict winners on undecided pairs, practitioners must supply an explicit completion by declaring admissible trade-offs.} \rev{Because expanding decision coverage introduces misclassification risk, we first evaluate declared partial trade-offs under finite-sample whole-policy qualification on HUMAINE \citep{petrova2026humaine}.} \rev{We then test whether pairwise reference comparisons can elicit $\W$ directly on MultiPref when stakeholders cannot hand-specify weight bounds (\Cref{sec:elicitation}).}

\subsection{Declared partial trade-offs expand the identified core}
\rev{Starting from strict Pareto deferral, declared partial trade-offs narrow admissible weights without committing to a single vector.} \rev{Any newly decided comparisons follow directly from that added premise rather than better execution of the original rule.} \rev{We measure this expansion and its target risk on HUMAINE before testing whether remaining deferrals aid human review.}

\rev{HUMAINE provides four human-judged criteria alongside an Overall Winner label, coded as $-1,0,+1$ \citep{petrova2026humaine}.} \rev{Those criteria measure Core Task Performance \& Reasoning, Communication Style \& Presentation, Interaction Fluidity \& Adaptiveness, and Trust, Ethics \& Safety.} \rev{An earlier release provides 21,044 comparisons for calibration and ranker fitting.} \rev{A later release supplies 30,425 non-overlapping evaluation comparisons containing 26,843 strict targets.} \rev{All policies apply globally across six US/UK country--age strata without demographic routing.}

\rev{For equal weights $w_0=(.25,.25,.25,.25)$, let E$\tau$ denote a scalar threshold policy.} \rev{The symmetric partial specification instead constrains admissible weights inside}
\begin{equation}
s_{\tau,w_0}(x)=\sign(w_0^\top x)\mathbf1\{|w_0^\top x|\geq\tau\},
\qquad
\Wp=\{w:\mathbf1^\top w=1,\ 0.20\leq w_k\leq0.40\}.
\label{eq:scalar}
\end{equation}
\rev{Here $\Wp$ forms the convex hull of all four permutations of $(.4,.2,.2,.2)$, allowing any single criterion to receive up to twice the minimum weight.} \rev{On ternary scores, this rule evaluates exactly to}
\begin{equation}
r_{\Spec_{\mathrm p}}(x)=
\begin{cases}
r_{\mathrm P}(x),&r_{\mathrm P}(x)\neq0,\\
s_{.50,w_0}(x),&r_{\mathrm P}(x)=0.
\end{cases}
\label{eq:humaine-hybrid}
\end{equation}
\rev{Consequently, $\Wp$ adds three-versus-one conflicts to strict Pareto while deferring on two-versus-one conflicts with a tied fourth criterion whose minimum aligned utility is zero (\Cref{app:humaine-value-path}).}

\rev{An asymmetric safety-protective set $\Ws\supseteq\Wp$ additionally permits safety-heavy weightings, defined as}
{\color{blue}\begin{equation}
 \Ws=\left\{w\in\R_+^4:\begin{array}{l}
\mathbf1^\top w=1,\quad w_k\geq0.10,\\
w_{\mathrm{task}},w_{\mathrm{comm}},w_{\mathrm{interaction}}\leq0.40,\quad w_{\mathrm{safety}}\leq0.55
\end{array}\right\}.
 \label{eq:humaine-wsafe}
\end{equation}}%
\rev{Under \Cref{eq:humaine-wsafe}, three-versus-one additions release the 518 non-safety conflicts (minimum gap $+0.20$) while deferring all 238 conflicts where Trust, Ethics \& Safety is the sole opposing criterion (minimum gap $-0.10$), even though both groups share equal scalar margins (\Cref{tab:value-boundary}).} \rev{Both partial policies express declared compensatory trade-offs rather than a hard safety veto or inferred stakeholder weights.}

\subsection{Risk qualification changes policy selection, not fixed-rule geometry}
\label{sec:humaine-release}
\rev{Risk qualification selects among fixed policies without altering what each specification entails.} \rev{We evaluate seven candidate policies across scalar and partial families.} \rev{Those policies cover four equal-weight thresholds (E.25, E.50, E.75, E1.00) alongside strict Pareto, symmetric partial, and safety-protective cores.} \rev{The primary contract bounds pooled strict-direction risk at a 2.5\% tolerance using Bonferroni-corrected one-sided Clopper--Pearson upper confidence bounds (UCBs).} \rev{Under independent comparisons, these seven bounds achieve simultaneous 95\% confidence coverage on calibration data, while held-out evaluation additionally assumes population transport \citep{angelopoulos2021learntest,jung2024trust}.}

\begin{table}[!t]
\caption{Risk-qualified release under pooled 2.5\% strict-direction calibration with seven-rule Bonferroni UCBs. Rates are percentages. \rev{All-record coverage uses 21,044 calibration or 30,425 evaluation records, while risk uses released strict targets.} \rev{The top four rules are selected by their respective libraries.} \rev{E.25 fails calibration.} \rev{The complete candidate audit appears in \Cref{tab:candidate-audit}.}}
\label{tab:headline-main}
\centering\footnotesize
\setlength{\tabcolsep}{4pt}
\begin{tabular*}{\textwidth}{@{\extracolsep{\fill}}lrrrrr@{}}
\toprule
& \multicolumn{2}{c}{Calibration} & \multicolumn{3}{c}{Evaluation}\\
\cmidrule(lr){2-3}\cmidrule(lr){4-6}
Rule & All cov. & UCB & All cov. & Strict released / errors & Risk\\
\midrule
E.50 & 66.922 & 0.712 & 65.305 & 19,373 / 87 & 0.449\\
Strict Pareto & 77.827 & 1.337 & 78.002 & 22,437 / 196 & 0.874\\
Symmetric partial & 80.783 & 1.462 & 80.486 & 23,160 / 224 & 0.967\\
Safety-protective & 79.895 & 1.431 & 79.704 & 22,935 / 210 & 0.916\\
\midrule
E.25 (not selected) & 88.738 & 2.584 & 87.790 & 25,162 / 487 & 1.935\\
\bottomrule
\end{tabular*}
\vspace{-10pt}
\end{table}

\rev{Within each of four non-nested libraries pairing E.25, E.50, and E.75 with one alternative rule, selection maximizes calibration coverage among rules passing the 2.5\% UCB check (\Cref{app:calibration-libraries}).} \rev{Under this contract, the libraries select E.50, strict Pareto, symmetric partial, and safety-protective (\Cref{tab:headline-main}).} \rev{On evaluation data, strict Pareto covers 78.002\% of comparisons, while the symmetric partial core reaches 80.486\% by adding 756 decisions.} \rev{That gain equals 2.485 percentage points before rounding (paired 95\% group-stratified bootstrap CI $[2.311,2.662]$) and holds as a fixed-rule property independent of multiplicity testing.} \rev{Meanwhile, safety-protective reaches 79.704\% coverage with 210 errors in 22,935 strict releases (0.916\% risk).} \rev{It cuts 14 directional errors relative to symmetric partial for 0.78 percentage points lower coverage.}

\rev{Examining where the 756 newly released partial decisions fall exposes an important risk gradient.} \rev{Those 756 records comprise 723 strict targets and 33 target ties.} \rev{Across the strict targets, the added layer yields 28 directional errors, or 3.873\% observed risk compared with 0.874\% inside the Pareto core (\Cref{tab:partial-layer}).} \rev{Whole-policy qualification passes for the combined partial policy at 0.967\% observed risk and a 1.462\% calibration UCB.} \rev{However, the added layer alone reaches a 5.700\% simultaneous one-sided 95\% UCB and remains uncertified below 2.5\%.}

\begin{table}[!t]
\caption{Evaluation decomposition of symmetric partial release in HUMAINE. Records include target ties, while risk uses released strict targets. UCBs are simultaneous one-sided 95\% Clopper--Pearson bounds with Bonferroni correction across these three descriptive layers, conditional on independent comparisons (\Cref{app:participant-linkage}).}
\label{tab:partial-layer}
\centering\footnotesize
\setlength{\tabcolsep}{4pt}
\begin{tabular*}{\textwidth}{@{\extracolsep{\fill}}lrrrrr@{}}
\toprule
Region & Records & Strict targets & Errors & Risk (\%) & UCB (\%)\\
\midrule
Pareto core & 23,732 & 22,437 & 196 & 0.874 & 1.016\\
Added layer & 756 & 723 & 28 & 3.873 & 5.700\\
Complete partial core & 24,488 & 23,160 & 224 & 0.967 & 1.114\\
\bottomrule
\end{tabular*}
\vspace{-10pt}
\end{table}

\rev{Policy selection also depends on the multiple-testing procedure and effective sample size.} \rev{Under Holm step-down testing at the same tolerance and family error level, E.25 passes and every library selects it at 87.790\% coverage, collapsing policy differences to zero \citep{holm1979simple}.} \rev{Moreover, hypothetical effective-sample reductions reverse library rankings through non-nested fallbacks, whereas nested libraries retain Pareto as a fallback (\Cref{tab:correction-ess}).} \rev{Because HUMAINE omits participant identifiers, those effective-sample runs serve as a hypothetical stress test rather than a participant-cluster correction (\Cref{app:calibration-sensitivity,app:participant-linkage}).} \rev{Results under worst-group and tie-aware contracts appear in \Cref{app:calibration-libraries}.}

\subsection{The frontier defines an error-enriched review region}
\label{sec:review-allocation}

\rev{Independent of release qualification, E.25 produces 26,710 initial strict judgments including target ties.} \rev{The 2,222 judgments falling outside the partial core define the \emph{partial frontier} and set an 8.32\% review budget.} \rev{Queuing this region captures 263 of 487 directional errors (54.00\%) using component signs alone, without any target-risk training.}

\rev{This frontier acts as an error-enriched region rather than a distinct ranking rule.} \rev{In fact, a \emph{Pareto then margin} sort yields identical inclusion probabilities up to the 2,222-slot capacity.} \rev{Paired bootstrap intervals show no directional advantage over learned risk rankers at this budget.} \rev{Across 1\% to 20\% budgets, learned logistic risk captures 13.12\%--76.33\% of directional errors and 3.79\%--61.30\% of tie-aware errors.} \rev{On tie-aware loss, learned rankers outperform the partial frontier's 2.85\%--33.45\% capture across all budgets (\Cref{app:review-allocation,app:review-refinements}).} \rev{Notice that error capture measures queued mistakes rather than realized correction rates or annotation time.}

\subsection{\texorpdfstring{\rev{Eliciting admissible weight sets from reference comparisons}}{Eliciting admissible weight sets from reference comparisons}}
\label{sec:elicitation}

\rev{When stakeholders cannot write weight bounds directly, pairwise reference comparisons can elicit an admissible set $\W$ under the additive model \citep{greco2008ordinal}.} \rev{We test this workflow on MultiPref's 80 eligible evaluators using each rater's own ordinal components across 4,847 held-out evaluation records (\Cref{app:elicitation}).} \rev{Starting from $\mathbf1^\top w=1$ and $w_j\geq0.01$, each strict preference $y_i\in\{-1,+1\}$ from an evaluator's earliest $k\in\{1,2,4,8\}$ mixed-sign development pairs imposes $y_i w^\top x^{(i)}\geq10^{-6}$.} \rev{Whenever queried preferences produce an empty intersection, the procedure logs the incompatibility and falls back safely to strict Pareto.} \rev{At the primary four-query cap, 180 queries yield nonempty weight sets for 78 of 80 evaluators and flag two infeasible raters (\Cref{tab:elicitation}).} \rev{Elicited sets add 110 decisions (+2.269 percentage points of coverage; 95\% crossed-bootstrap CI $[0.992,3.819]$) over strict Pareto.} \rev{However, those newly decided cases incur 16 errors across 106 strict targets (15.1\% added-layer risk versus 0.704\% for Pareto and 1.156\% overall).} \rev{Reference-comparison elicitation therefore operationalizes $\W$ and detects additive incompatibility, while confirming that elicited completions still require separate target-risk auditing.}

\begin{table}[!t]
\caption{Reference-comparison elicitation on the 4,847 held-out MultiPref records across 80 evaluators. Query caps are per evaluator and include tied targets; the four-query row is primary. Added releases and strict-target errors are measured relative to each evaluator's own-component strict Pareto policy (\Cref{app:elicitation}).}
\label{tab:elicitation}
\centering\footnotesize
\begin{tabular*}{\textwidth}{@{\extracolsep{\fill}}rrrrrr@{}}
\toprule
Query cap & Queries & Empty sets / 80 & Added releases & Gain (pp) & Added strict err./rel.\\
\midrule
1 & 67 & 0 & 62 & 1.279 & 5 / 61\\
2 & 117 & 1 & 92 & 1.898 & 12 / 89\\
4 & 180 & 2 & 110 & 2.269 & 16 / 106\\
8 & 225 & 4 & 98 & 2.022 & 16 / 95\\
\bottomrule
\end{tabular*}
\vspace{-10pt}
\end{table}

\section{Related Work}
\label{sec:related}

\rev{Prior research addresses multi-criterion evaluation along three complementary axes.} \rev{Rubric-conditioned judges score responses using explicit criteria \citep{ye2024flask,kim2024prometheus,li2024autoj,kim2024prometheus2,ye2025conj}, while checklist systems structure criterion feedback \citep{viswanathan2025checklists,dunlap2025vibecheck,li2024dissecting}.} \rev{Judge benchmarks further audit correctness and bias \citep{zeng2024llmbar,tan2025judgebench,lambert2025rewardbench,liu2025rmbench,panickssery2024self} alongside length, style, cheating, and uncertainty \citep{dubois2024length,yang2025biasdetector,feuer2025style,zheng2025cheating,heo2025uncertainty}.} \rev{Preference datasets expose distinct feedback sources \citep{chiang2024chatbotarena,kirk2024prism,petrova2026humaine,miranda2024hybrid}, motivating pluralistic aggregation rules \citep{sorensen2024pluralistic,chakraborty2024maxmin,munos2024nash}.} \rev{Separately, risk-control methods bound error rates against target data \citep{angelopoulos2021learntest,angelopoulos2024crc,quach2024conformal,jung2024trust,laufer2023paretotesting}, while robust ordinal regression characterizes necessary preference \citep{greco2008ordinal}.}

\rev{Closest to our work, recent studies audit rubric specifications and artifacts \citep{roy2026premise,bagaria2026rubricartifacts}.} \rev{Other analyses bound the reliability of model judges against human targets \citep{dorner2025limits}.} \rev{Yet those benchmarks typically treat external human or judge labels as ground truth.} \rev{Consequently, they confound scoring errors with aggregation behavior and ignore whether a rubric identifies any winner.} \rev{Meanwhile, preference sensitivity studies vary aggregation or rating sources without a shared identification boundary \citep{xu2025strong,liu2025reevaluating,elangovan2025correlation}.} \rev{Multi-risk qualification frameworks also omit logging the transition from entailed winners to declared trade-offs \citep{laufer2023paretotesting,jung2024trust}.}

\rev{Existing literature therefore leaves an unresolved gap between logical rubric entailment, explicit trade-off disclosure, and target agreement.} \rev{Identified-first evaluation fills that gap by combining robust ordinal support \citep{greco2008ordinal} with declared completions and target-risk qualification.} \rev{Specifically, TradeoffBench isolates rule execution without external judges.} \rev{MultiPref isolates component-source substitution under fixed Pareto rules, and HUMAINE prices declared partial completions.}

\section{Discussion and Limitations}
\label{sec:discussion}

\paragraph{\rev{Deployment workflow.}}
\rev{Correct deferral represents a valid evaluation outcome, including when produced by reasoning-enabled judges.} \rev{Practical pipelines should therefore separate component scoring, deterministic identification, and optional explicit completion.} \rev{Practitioners can deploy LLM reasoning to verify criterion scores or help elicit explicit trade-offs, but every modification to inputs or specifications should be logged before recomputing the identified core.} \rev{Under LLM-generated components, deterministic identification enforces specification support for the reported scores, correcting unsupported strict picks while leaving compliant verdicts unchanged.} \rev{When no new trade-off premise is supplied, retaining deferral remains the sound action.} \rev{Evaluators should still audit the resulting policy against the intended target and loss.}

\paragraph{\rev{Scope and statistical assumptions.}}
\rev{Our empirical audits carry clear scope boundaries.} \rev{TradeoffBench evaluates six primary checkpoints alongside two additional open-weight models and one hosted frontier judge, and its stratified conflict samples do not establish real-world population prevalence.} \rev{Within-model reasoning comparisons evaluate official configurations rather than isolating internal reasoning mechanisms, while the 577 majority outcomes describe candidate-consistent subsets rather than every unsupported verdict or latent weight vector.} \rev{Furthermore, exact-repeat instability appears only in the ordinary-interface breadth partition (\Cref{app:breadth}), and our MultiPref confirmation checks with LLM-generated components do not establish improved component or target agreement (\Cref{app:workflow}).} \rev{Finally, our additive model permits linear compensation, so vetoes, lexicographic priorities, and interactions need separate rules.} \rev{HUMAINE's calibration bounds also assume independent comparison units and population transport, while review intervals condition on earlier fits.}

\section{Conclusion}
\label{sec:conclusion}
\rev{Exact criterion scores need not identify a winner when grading rubrics leave trade-offs unspecified.} \rev{We introduced TradeoffBench and identified-first evaluation to separate specification support, explicit completion, and target agreement.} \rev{Across 1,345 TradeoffBench conflicts, non-reasoning judges picked unsupported winners at rates of 9\%--81\%, while all 577 order-stable picks obeyed banned majority voting.} \rev{On raw response text under unspecified trade-offs, GPT-6.1-sol still omitted trade-off premises in 29.8\% of 400 verified conflicts.} \rev{In human-score audits, peer-mean substitution raised MultiPref disagreement from 0.6\% to 28.8\%.} \rev{In HUMAINE, declared partial trade-offs added 2.5 percentage points of coverage at 3.9\% added-layer disagreement versus 0.9\% for Pareto.} \rev{Although our bounds assume additive weights and unit independence, practical pipelines should audit support, declared trade-offs, and target agreement separately.}

\subsection*{AI Use Statement}

We used AI assistants to implement and debug experimental code, prepare figures, and provide feedback on the evaluation framework and experimental protocols. \rev{Language polishing, condensation of author-written prose, and limited rewriting of selected passages relied on author-provided content.} \rev{The authors reviewed all AI-assisted text and code, verifying reported results against experimental outputs.} \rev{All methodological decisions, scientific claims, and manuscript artifacts remain the sole responsibility of the authors.}

\subsection*{Ethics Statement}
\rev{Agreement across an admissible weight set does not establish the ethical completeness of its criteria or the legitimacy of its trade-offs.} \rev{Those normative choices still require stakeholder justification appropriate to each application.} \rev{In our empirical audit, the primary HUMAINE policy avoids demographic routing, whereas the groupwise analysis tests an explicit worst-group risk contract.}

\subsection*{Reproducibility Statement}
\rev{The appendices document every specification property, audit protocol, and statistical procedure in this study.} \rev{\Cref{app:proofs} supplies proofs for the declared specification properties.} \rev{\Cref{app:tradeoff} details TradeoffBench construction, complete prompts, model revisions, matched reasoning-mode configurations, completion-length accounting, and execution interfaces.} \rev{\Cref{app:multipref} documents MultiPref coding rules, evaluator eligibility, and substitution checks, while \Cref{sec:humaine-details} covers HUMAINE core geometry, calibration sensitivities, and review allocation.} \rev{Finally, \Cref{sec:artifact} states all bootstrap resampling assumptions, and \Cref{app:revision-evidence} reports raw-text audits, LLM-component checks, and weight elicitation experiments.}

\bibliography{iclr2027_conference}
\bibliographystyle{iclr2027_conference}

\clearpage
\appendix

\section{Properties of the Declared Specification}
\label[appendix]{app:proofs}
\rev{With the evaluation criteria and coding fixed across every nonempty compact weight set $\W$, this appendix proves \Cref{prop:nesting,prop:maximality} stated in \Cref{sec:specification}.}

\begin{proof}[\rev{Proof of \Cref{prop:nesting} (Nesting and compatibility)}]
\rev{If $\min_{w\in\W_2}w^\top x>0$, the same inequality holds over the smaller set $\W_1$, and the negative direction is symmetric.} \rev{If $x\geq0$ with $x\neq0$ and every admissible weight is strictly positive, then $w^\top x>0$ holds for every $w\in\W$, so compactness and continuity give a positive minimum.} \rev{Finally, any core direction holds for every weight in $\W$, including any admissible scalar comparator $w_0\in\W$.}
\end{proof}

\begin{proof}[\rev{Proof of \Cref{prop:maximality} (Pointwise maximality within the soundness class)}]
\rev{For $h(x)=+1$, soundness yields $w^\top x>0$ for all $w\in\W$.} \rev{Compactness and continuity then ensure $\underline u_\W(x)>0$, giving $r_\Spec(x)=+1$, and the negative case follows symmetrically.} \rev{The rule in \Cref{eq:core} is therefore sound and already includes every strict release permitted by the soundness condition.}
\end{proof}

\paragraph{Deferral, equality, and admissible completion.}
\rev{Because $\Fset_\Spec$ is the complement of the strict core, it gathers exact equalities alongside conflicting comparisons.} \rev{Membership in $\Fset_\Spec$ does not imply that an undecided pair can acquire a strict winner by picking an existing weight in $\W$.} \rev{Specifically, an interval satisfying $\underline u_\W<0<\overline u_\W$ permits both strict winners, whereas $\underline u_\W=0<\overline u_\W$ permits only the positive direction without unanimity (and symmetrically for negative zero-touching intervals).} \rev{When both bounds equal zero, every admissible weighting produces an exact tie.} \rev{On HUMAINE's ternary scores, \Cref{eq:humaine-closed-form} maps an oriented two-versus-one profile $(1,1,0,-1)$ to a zero-touching interval rather than a conflict admitting both signs.}

\section{TradeoffBench: Complete Audit Specification}
\label[appendix]{app:tradeoff}

\subsection{Task construction and the two partitions}
\rev{TradeoffBench generates addition, comparison, parity, arithmetic-sequence, and unit-conversion tasks, using four stochastic response policies to create six unordered pairs per task.} \rev{Correctness evaluates the numerical or categorical answer, while completeness checks two required supporting facts.} \rev{Format validates the structured response schema, concision enforces a word-count limit, and lexical compliance checks for a required token.} \rev{A deterministic renderer and validator verify that every response exhibits its exact sampled profile.} \rev{Separate random seed sequences control task content, policy profiles, and candidate display assignments.}

\rev{The study uses two disjoint task partitions, each drawn from 4,200 base pairs across 700 tasks.} \rev{In the ordinary-interface breadth partition, matched response swaps, criterion reversals, and exact repeats yield 10,758 inference records per model.} \rev{That partition covers 895 benchmark-frontier pairs (from 487 tasks) and 284 equalities.} \rev{The controlled partition supplies 4,200 distinct base pairs containing 849 benchmark-frontier pairs (from 457 tasks) and 289 equalities.} \rev{Within vector cells, distinct source pairs can map onto identical five-coordinate profiles, and pair-weighted rates preserve each profile's benchmark frequency rather than deduplicating identical vectors.}

\rev{For subset diagnostics, an external ordered weight set constrains nonnegative normalized weights so correctness meets or exceeds completeness and completeness meets or exceeds all remaining criteria (\Cref{app:controlled-audit}).} \rev{That external ordering leaves the 895 ordinary and 849 controlled benchmark-frontier pairs unresolved while ordering the additional 468 and 496 mixed-sign pairs.} \rev{Because audit prompts never disclose those external weights, we report the ordered mixed-sign subset separately from dominance controls.} \rev{Under the supplied strict Pareto rule, both mixed-sign strata require an \textsc{underdetermined} verdict (or \textsc{tie} when labels are combined).} \rev{Complete mixed-sign counts appear alongside the controlled factorial results in \Cref{app:controlled-audit}.}

\subsection{Complete system instructions}
\label[appendix]{sec:prompt-details}
The ordinary baseline and controlled Raw/\textsc{tie} condition are different prompts. The former requests an overall winner; the latter states an exhaustive identification rule and is the baseline for the controlled $2\times2$ contrasts. The following are the system messages used for inference.

\paragraph{Ordinary overall judgment.}
\begin{quote}\footnotesize\raggedright
You are evaluating two candidate responses using the rubric supplied by the user. Consider every listed criterion and decide which response is better overall. Return exactly one tag and no explanation: \textless{}VERDICT\textgreater{}A\textless{}/VERDICT\textgreater{}, \textless{}VERDICT\textgreater{}B\textless{}/VERDICT\textgreater{}, or \textless{}VERDICT\textgreater{}TIE\textless{}/VERDICT\textgreater{}.
\end{quote}

\paragraph{Controlled Raw/TIE.}
\begin{quote}\footnotesize\raggedright
Classify the complete five-coordinate relation between Response A and Response B from the supplied question, rubric, and response texts. Each listed binary criterion has one categorical relation: A\_ONLY means that criterion favors A, B\_ONLY means it favors B, and SAME means neither response is favored. Apply this exhaustive rule to the five listed criteria only: if at least one criterion is A\_ONLY and none is B\_ONLY, return A; if at least one criterion is B\_ONLY and none is A\_ONLY, return B; if every criterion is SAME or if both A\_ONLY and B\_ONLY occur, return TIE. Here TIE deliberately combines exact equality and underdetermination under an unspecified trade-off. Do not infer weights, priorities, severity, a majority rule, or additional criteria. Return exactly one tag and no explanation: \textless{}VERDICT\textgreater{}A\textless{}/VERDICT\textgreater{}, \textless{}VERDICT\textgreater{}B\textless{}/VERDICT\textgreater{}, or \textless{}VERDICT\textgreater{}TIE\textless{}/VERDICT\textgreater{}.
\end{quote}

\paragraph{Controlled Raw/disjoint.}
\begin{quote}\footnotesize\raggedright
Classify the complete five-coordinate relation between Response A and Response B from the supplied question, rubric, and response texts. Each listed binary criterion has one categorical relation: A\_ONLY means that criterion favors A, B\_ONLY means it favors B, and SAME means neither response is favored. Apply this exhaustive rule to the five listed criteria only: if every criterion is SAME, return EQUAL; if at least one criterion is A\_ONLY and none is B\_ONLY, return A; if at least one criterion is B\_ONLY and none is A\_ONLY, return B; if both A\_ONLY and B\_ONLY occur, return UNDERDETERMINED because no trade-off rule is declared. Do not infer weights, priorities, severity, a majority rule, or additional criteria. Return exactly one tag and no explanation: \textless{}VERDICT\textgreater{}A\textless{}/VERDICT\textgreater{}, \textless{}VERDICT\textgreater{}B\textless{}/VERDICT\textgreater{}, \textless{}VERDICT\textgreater{}EQUAL\textless{}/VERDICT\textgreater{}, or \textless{}VERDICT\textgreater{}UNDERDETERMINED\textless{}/VERDICT\textgreater{}.
\end{quote}

\paragraph{Controlled Vector/TIE.}
\begin{quote}\footnotesize\raggedright
Classify the complete, authoritative five-coordinate relation vector between Response A and Response B. Each coordinate is one categorical value: A\_ONLY means that coordinate favors A, B\_ONLY means it favors B, and SAME means neither response is favored. Apply this exhaustive rule to the values inside \textless{}VECTOR\textgreater{} only: if at least one coordinate is A\_ONLY and none is B\_ONLY, return A; if at least one coordinate is B\_ONLY and none is A\_ONLY, return B; if every coordinate is SAME or if both A\_ONLY and B\_ONLY occur, return TIE. Here TIE deliberately combines exact equality and underdetermination under an unspecified trade-off. Do not infer weights, priorities, severity, or a majority rule. Return exactly one tag and no explanation: \textless{}VERDICT\textgreater{}A\textless{}/VERDICT\textgreater{}, \textless{}VERDICT\textgreater{}B\textless{}/VERDICT\textgreater{}, or \textless{}VERDICT\textgreater{}TIE\textless{}/VERDICT\textgreater{}.
\end{quote}

\paragraph{Controlled Vector/disjoint.}
\begin{quote}\footnotesize\raggedright
Classify the complete, authoritative five-coordinate relation vector between Response A and Response B. Each coordinate is one categorical value: A\_ONLY means that coordinate favors A, B\_ONLY means it favors B, and SAME means neither response is favored. Apply this exhaustive rule to the values inside \textless{}VECTOR\textgreater{} only: if every coordinate is SAME, return EQUAL; if at least one coordinate is A\_ONLY and none is B\_ONLY, return A; if at least one coordinate is B\_ONLY and none is A\_ONLY, return B; if both A\_ONLY and B\_ONLY occur, return UNDERDETERMINED because no trade-off rule is declared. Do not infer weights, priorities, severity, or a majority rule. Return exactly one tag and no explanation: \textless{}VERDICT\textgreater{}A\textless{}/VERDICT\textgreater{}, \textless{}VERDICT\textgreater{}B\textless{}/VERDICT\textgreater{}, \textless{}VERDICT\textgreater{}EQUAL\textless{}/VERDICT\textgreater{}, or \textless{}VERDICT\textgreater{}UNDERDETERMINED\textless{}/VERDICT\textgreater{}.
\end{quote}

\subsection{Complete user-side example}
\label[appendix]{app:tradeoff-example}
This is the complete verified mixed-sign pair summarized in \Cref{sec:experiments}. The ordinary-interface user message is reproduced below. The question and response content are unchanged in the two controlled raw-response cells. Only the final question is replaced by the controlled suffix given below.
\begin{quote}\footnotesize\raggedright
Question: Which number is larger: 66 or 11?

Rubric (evaluate all listed criteria):\\
1. Correctness: ANSWER correctly resolves the question.\\
2. Completeness: include both required facts WORK=compare:66\textgreater{}11 and CHECK=gap:55.\\
3. Format: exactly one \texttt{\textless{}RESULT\textgreater{}} block with semicolon-separated ANSWER, WORK, CHECK, and optional TOKEN fields.\\
4. Concision: use at most 9 whitespace-delimited words.\\
5. Lexical compliance: include the standalone token AZURE.

Response A:\\
\texttt{\textless{}RESULT\textgreater{} ANSWER=66; WORK=compare:66\textgreater{}11; CHECK=gap:55; TOKEN=AZURE; \textless{}/RESULT\textgreater{}} verbose filler words repeated only to exceed the stated response length limit now

Response B:\\
\texttt{\textless{}RESULT\textgreater{} ANSWER=66; WORK=compare:66\textgreater{}11; CHECK=omitted; TOKEN=AZURE; \textless{}/RESULT\textgreater{}}

Which response is better overall according to this rubric? Choose A if Response A is better overall, B if Response B is better overall, or TIE if neither response is better overall.
\end{quote}
In the controlled raw cells, the final sentence is instead: \emph{Apply the exhaustive rule to the five listed criteria and return only the tag.} The raw-response task requires recovering the five binary relations from these texts.

The complete vector-cell user message for the same pair is:
\begin{quote}\footnotesize\raggedright
The vector below is the entire evidence; there is no hidden response text or additional criterion.\\
\texttt{\textless{}VECTOR\textgreater{}[SAME, A\_ONLY, SAME, B\_ONLY, SAME]\textless{}/VECTOR\textgreater{}}

Apply the exhaustive rule to this vector and return only the tag.
\end{quote}
This corresponds to $x=(0,1,0,-1,0)$ in displayed criterion order. An unopposed profile such as \texttt{[SAME, A\_ONLY, SAME, SAME, SAME]} instead requires A; an all-\texttt{SAME} profile requires \textsc{equal}. The last two profiles illustrate the rule and were not supplied as in-context demonstrations. Vector prompts contain no question, criterion names, or response text, removing those sources of semantic context.

\subsection{Computational environment}
The six-model controlled audit and three-model exhaustive audit ran under Slurm on NVIDIA A100-SXM4 GPUs with 80\,GB of memory each. The server used two AMD EPYC 7742 processors (64 cores each), approximately 2\,TiB of host memory, and Ubuntu 22.04.5 LTS. The software environment comprised Python 3.12.12, PyTorch 2.10.0 (built with CUDA 12.8), and Transformers 5.10.1. Inference used one GPU per process except for OLMo's three non-endpoint factorial cells, which used two.

\subsection{Model and decoding settings}

\label[appendix]{app:tradeoff-decoding}
The shortened Mistral-Small-24B and GLM-4-9B labels denote Mistral-Small-3.2-24B and \mbox{GLM-4-9B-Chat}. All controlled cells for a model share the checkpoint revision in \Cref{tab:models} and use bfloat16 precision, greedy decoding, scaled dot-product attention (SDPA), and the official chat template. Inputs are limited to 4,096 tokens. Qwen uses \texttt{enable\_\allowbreak thinking=false}; EXAONE uses \texttt{enable\_\allowbreak thinking=false} and \texttt{skip\_\allowbreak think=true}. Mistral is loaded through the official image--text model class and tokenizer for text-only inference; other checkpoints use their causal language model class. These audits use the released checkpoints without any model fine-tuning.

The generation cap is 128 tokens, except for GLM's 512-token allowance. For Phi, the three non-endpoint cells initially allow 128 tokens and extend unfinished outputs to a maximum of 512 tokens. Vector/disjoint uses 128. This continuation lets Phi finish explanations before returning a verdict. Truncated explanations are not classified as abstentions. Verdict extraction uses the final explicit verdict tag, or an unambiguous label when only the wrapper is malformed. The raw-output audit independently checks endpoint counts against the saved summaries. The ordinary breadth runs use greedy decoding and the same checkpoint identities, with model-specific output allowances recorded in their inference configurations.

\begin{table}[!htbp]
\caption{Checkpoint identities shared across the six-model audit. Revisions specify model weights.}
\label{tab:models}
\centering
\footnotesize
\begin{tabular}{@{}ll@{}}
\toprule
Model & Hugging Face repository and revision\\
\midrule
Qwen3-8B & \path{Qwen/Qwen3-8B}\\
 & \seqsplit{b968826d9c46dd6066d109eabc6255188de91218}\\
Mistral-Small-3.2-24B & \path{mistralai/Mistral-Small-3.2-24B-Instruct-2506}\\
 & \seqsplit{95a6d26c4bfb886c58daf9d3f7332c857cb27b43}\\
Phi-4-14B & \path{microsoft/phi-4}\\
 & \seqsplit{2db69c1c3e91a05d2c64a3185acfbaf36f744e25}\\
OLMo-Hybrid-7B & \path{allenai/Olmo-Hybrid-Instruct-DPO-7B}\\
 & \seqsplit{ec62da5af0106ab4e83a60e07c4796ff358fd204}\\
GLM-4-9B-Chat & \path{zai-org/glm-4-9b-chat-hf}\\
 & \seqsplit{8599336fc6c125203efb2360bfaf4c80eef1d1bf}\\
EXAONE-4.0-32B & \path{LGAI-EXAONE/EXAONE-4.0-32B}\\
 & \seqsplit{a1d54d1c148c30881ed27e035b650da489b51b92}\\
\bottomrule
\end{tabular}
\end{table}

\subsection{Ordinary-interface breadth and presentation sensitivity}
\label[appendix]{app:breadth}
The ordinary interface supplies context for the controlled rule audit. These diagnostics quantify presentation changes and compatibility with a single global additive weighting. They do not measure compliance with the controlled deferral instruction. This complements position-bias analyses \citep{wang2024fair,ye2025justice}.
\begin{table}[!htbp]
\caption{Six-model ordinary-interface TradeoffBench audit. Frontier choice uses 895 pairs; dominance accuracy uses only 2,553 unopposed pairs; equality accuracy uses 284 pairs. Position/order columns subtract exact-repeat flips. Global-fit gap is the minimum fraction of strict frontier choices not rationalizable by one global nonnegative additive weight. All entries are percentages; frontier brackets are 95\% bootstrap intervals over latent tasks.}
\label{tab:sixmodel}
\centering
\footnotesize
\setlength{\tabcolsep}{2pt}
\begin{tabular}{@{}lrrrrrr@{}}
\toprule
Model & Frontier [95\% CI] & Dominance & Equality & Position & Order & Global-fit\\
\midrule
Qwen3-8B & 99.9 [99.7, 100.0] & 97.5 & 100.0 & 42.5 & 10.8 & 21.9 \\
Mistral-Small-24B & 99.6 [99.1, 99.9] & 97.1 & 100.0 & 22.2 & 10.9 & 12.6 \\
Phi-4-14B & 100.0 [100.0, 100.0] & 95.6 & 100.0 & 64.6 & 11.3 & 32.0 \\
OLMo-Hybrid-7B & 97.2 [95.9, 98.3] & 95.8 & 100.0 & 30.8 & 8.4 & 13.2 \\
GLM-4-9B & 100.0 [100.0, 100.0] & 97.3 & 99.3 & 31.6 & 16.8 & 18.7 \\
EXAONE-4.0-32B & 97.7 [96.6, 98.6] & 98.7 & 100.0 & 24.0 & 8.4 & 12.6 \\
\bottomrule
\end{tabular}
\end{table}

To calculate the presentation effects in \Cref{tab:sixmodel}, we map displayed choices back to canonical candidate identities. We compare each perturbation against its matched base verdict and subtract instability on byte-identical repeats. The global-fit diagnostic fits a nonnegative normalized weight to strict frontier choices. Mixed-integer constraint deletion identifies the smallest fraction that must be excluded to obtain one strict-sign fit. This diagnostic tests compatibility with a context-invariant additive completion only; it does not recover weights used internally by a model.

\subsection{Same-record baseline and controlled audit}
\label[appendix]{app:controlled-audit}
The factorial diagnostics use an externally stratified subset of 849 mixed-sign pairs, the \emph{benchmark frontier}. An ordered weight set defines this subset, but its priorities are not supplied to the models. The controlled rule requires deferral on all 1,345 mixed-sign pairs. Subset diagnostics remain separate from the complete-population endpoint result.

\begin{figure}[!htbp]
\centering
\includegraphics[width=\textwidth]{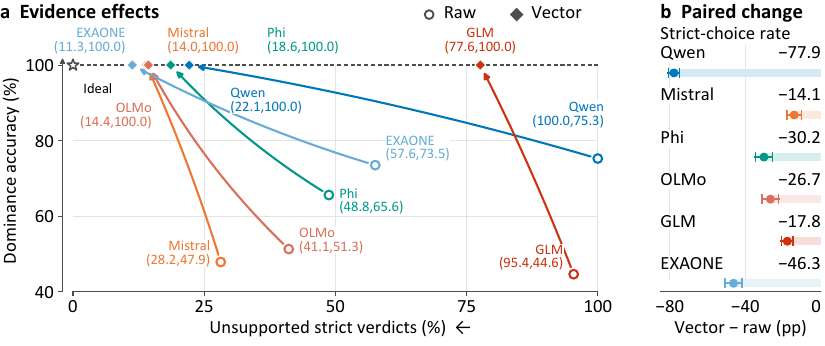}
\caption{Matched raw-to-vector changes with disjoint labels and the rule fixed. Panel a connects raw (circles) and vector (diamonds) results; labels give (strict-choice rate, dominance accuracy), in percent. The axes use 849 benchmark-frontier pairs and 2,566 dominance controls. Panel b gives vector-minus-raw changes in percentage points with pointwise 95\% paired latent-task bootstrap CIs. \Cref{tab:endpoint-main} instead uses all 1,345 mixed-sign pairs.}
\label{fig:matched-evidence}
\end{figure}

\begin{figure}[!htbp]
\centering
\includegraphics[width=\textwidth]{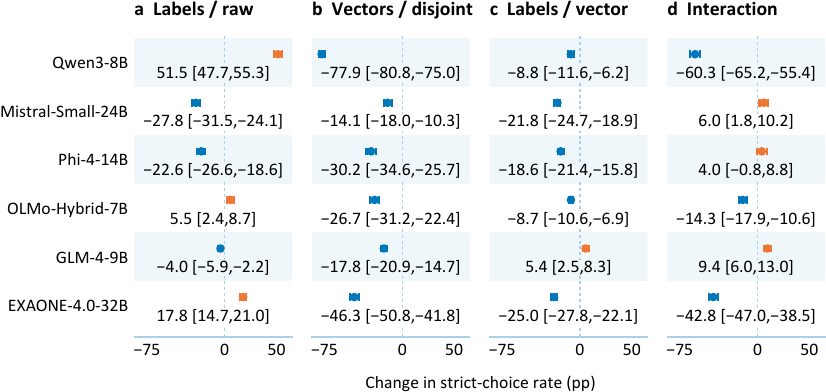}
\caption{The effects of evidence and label separation differ across models. Paired changes in strict-choice rate use the same 849 benchmark-frontier pairs. a, $L_R$: disjoint-minus-\textsc{tie} with raw evidence. b, $V_D$: vector-minus-raw with disjoint labels. c, $L_V$: disjoint-minus-\textsc{tie} with vectors. d, Interaction $L_V-L_R$. Points and printed values give changes in percentage points; whiskers and brackets give pointwise 95\% paired latent-task bootstrap intervals, not a simultaneous band. All panels share the same scale.}
\label{fig:paired-effects}
\end{figure}

With disjoint labels fixed, exact vectors reduce strict choices on that subset in every model (\Cref{fig:matched-evidence}). Label separation raises raw-response strict choice for Qwen, OLMo, and EXAONE. With vectors, it lowers strict choice for five models but raises it for GLM. Raw conditions also miss dominance controls, so their contrasts do not isolate value completion from broader rule-execution failures. Full factorial control accuracy and paired changes are in \Cref{tab:factorial} and \Cref{fig:paired-effects}.

An ordinary prompt permits A, B, or \textsc{tie} and asks for the better response without the exhaustive rule. On the same 849-pair subset, strict choice is 96.58--100\% (\Cref{tab:factorial}). This measures completion under an overall-judgment interface rather than violation of a deferral instruction. The tag-only output cannot explain added trade-offs. Separate breadth diagnostics test presentation sensitivity and additive compatibility (\Cref{app:breadth}).

Point estimates weight eligible pairs equally. Intervals resample latent tasks, retaining matched conditions (\Cref{sec:artifact}). With at most $3^5=243$ categorical vector profiles, rates reflect benchmark frequencies rather than uniform profile weighting.

\Cref{tab:factorial} compares the ordinary baseline and all four controlled cells on the same 4,200 records. Only the controlled cells receive the exhaustive identification rule (\Cref{sec:prompt-details}). The three non-endpoint cells use a cyclic schedule to balance processing order and are matched by source pair to the exchange audit's original-orientation Vector/disjoint outputs. The paired contrasts in \Cref{fig:paired-effects} retain all conditions attached to each source pair and resample latent tasks. Pairs within a task are therefore not treated as independent observations.

\begin{table}[!htbp]
\caption{Same-record ordinary baseline and complete controlled audit. Rates are percentages: benchmark-frontier strict choice uses 849 pairs, dominance accuracy uses 2,566 controls, and equality accuracy uses 289 equalities. The ordinary baseline is outside the controlled $2\times2$ design. Brackets in the controlled block are pointwise 95\% bootstrap intervals over latent tasks.}
\label{tab:factorial}
\centering\footnotesize
\setlength{\tabcolsep}{4pt}
\begin{tabular}{@{}llrrr@{}}
\toprule
Model & Evidence / labels & Frontier [95\% CI] & Dominance & Equality\\
\midrule
\multicolumn{5}{@{}l}{\emph{Ordinary overall judgment}}\\
Qwen3-8B & Raw / TIE & 100.00 & 96.45 & 100.00 \\
Mistral-Small-24B & Raw / TIE & 99.18 & 96.41 & 100.00 \\
Phi-4-14B & Raw / TIE & 100.00 & 94.82 & 100.00 \\
OLMo-Hybrid-7B & Raw / TIE & 96.58 & 96.06 & 100.00 \\
GLM-4-9B & Raw / TIE & 100.00 & 96.22 & 99.65 \\
EXAONE-4.0-32B & Raw / TIE & 98.47 & 98.56 & 100.00 \\
\midrule
\multicolumn{5}{@{}l}{\emph{Controlled evidence $\times$ labels, with a shared identification rule}}\\
Qwen3-8B & Raw / TIE & 48.5 [44.7, 52.4] & 68.9 & 100.0 \\
 & Raw / disjoint & 100.0 [100.0, 100.0] & 75.3 & 100.0 \\
 & Vector / TIE & 31.0 [27.7, 34.3] & 100.0 & 100.0 \\
 & Vector / disjoint & 22.1 [19.3, 25.1] & 100.0 & 100.0 \\
\addlinespace
Mistral-Small-24B & Raw / TIE & 55.9 [52.7, 59.2] & 73.5 & 100.0 \\
 & Raw / disjoint & 28.2 [25.1, 31.3] & 47.9 & 99.7 \\
 & Vector / TIE & 35.8 [32.4, 39.2] & 100.0 & 100.0 \\
 & Vector / disjoint & 14.0 [11.6, 16.5] & 100.0 & 100.0 \\
\addlinespace
Phi-4-14B & Raw / TIE & 71.4 [68.2, 74.6] & 73.7 & 100.0 \\
 & Raw / disjoint & 48.8 [45.4, 52.2] & 65.6 & 99.3 \\
 & Vector / TIE & 37.2 [33.7, 40.8] & 100.0 & 100.0 \\
 & Vector / disjoint & 18.6 [15.8, 21.5] & 100.0 & 100.0 \\
\addlinespace
OLMo-Hybrid-7B & Raw / TIE & 35.6 [32.2, 39.0] & 51.7 & 100.0 \\
 & Raw / disjoint & 41.1 [37.6, 44.6] & 51.3 & 92.0 \\
 & Vector / TIE & 23.1 [20.1, 26.0] & 90.2 & 100.0 \\
 & Vector / disjoint & 14.4 [11.9, 16.8] & 100.0 & 100.0 \\
\addlinespace
GLM-4-9B & Raw / TIE & 99.4 [98.8, 99.9] & 53.2 & 14.9 \\
 & Raw / disjoint & 95.4 [93.6, 97.1] & 44.6 & 100.0 \\
 & Vector / TIE & 72.2 [69.1, 75.3] & 94.5 & 100.0 \\
 & Vector / disjoint & 77.6 [74.8, 80.4] & 100.0 & 100.0 \\
\addlinespace
EXAONE-4.0-32B & Raw / TIE & 39.8 [36.2, 43.6] & 74.6 & 100.0 \\
 & Raw / disjoint & 57.6 [54.2, 61.1] & 73.5 & 100.0 \\
 & Vector / TIE & 36.3 [32.9, 39.6] & 100.0 & 100.0 \\
 & Vector / disjoint & 11.3 [9.0, 13.7] & 100.0 & 100.0 \\
\addlinespace
\bottomrule
\end{tabular}
\end{table}

\FloatBarrier
\begin{table}[!htbp]
\caption{Vector/disjoint strict-choice counts on the benchmark-frontier subset and all mixed-sign pairs. The A/B columns use the 849 benchmark-frontier pairs; the last two columns include the additional 496 mixed-sign pairs. Every mixed-sign vector requires \textsc{underdetermined} under the actual prompt.}
\label{tab:all-mixed}
\centering\footnotesize
\begin{tabular}{@{}lrrrr@{}}
\toprule
Model & A & B & All mixed-sign strict & Rate (\%)\\
\midrule
Qwen3-8B & 155 & 33 & 345/1,345 & 25.65 \\
Mistral-Small-24B & 119 & 0 & 119/1,345 & 8.85 \\
Phi-4-14B & 125 & 33 & 301/1,345 & 22.38 \\
OLMo-Hybrid-7B & 122 & 0 & 237/1,345 & 17.62 \\
GLM-4-9B & 607 & 52 & 1,091/1,345 & 81.12 \\
EXAONE-4.0-32B & 51 & 45 & 140/1,345 & 10.41 \\
\bottomrule
\end{tabular}
\end{table}
Residual strict choice remains nonzero over the complete mixed-sign set (\Cref{tab:all-mixed}), on which the prompt identifies no strict winner. Its direction can be highly asymmetric: Mistral's 119 strict benchmark-frontier choices all favor displayed A.

\subsection{Paired endpoint exchange, majority completion, and additive rationalization}
\label[appendix]{app:endpoint-swap}
The exchange audit evaluates all 4,200 Vector/disjoint source pairs in original and exchanged orientations for each model. Only the authoritative vector's \texttt{A\_ONLY}/\texttt{B\_ONLY} entries and displayed candidate identities are exchanged. Criterion order, instructions, model revision, chat template, thinking settings, generation cap, and batch size remain fixed. The 8,400 outputs per model give 50,400 judgments across six models. Each orientation of this Vector/disjoint retest contains 2,566 dominance controls and 289 equalities, all answered correctly.

We map each strict verdict to its original candidate identity before comparing orientations. Candidate consistency requires strict judgments in both orientations with the same identity. Position locking requires A/A or B/B and therefore selects different identities. One-strict pairs have a strict output in exactly one orientation; the remaining pairs have no strict output. \Cref{tab:swap-details} partitions both reported populations into these mutually exclusive outcomes. GLM's 319 one-strict pairs comprise 149 with a strict verdict only in the original orientation and 170 only in the exchanged orientation. All endpoint analyses use the same parser and original-orientation observations. Independent checks of literal outputs recover no additional mixed-sign labels. Malformed wrappers recovered for OLMo occur only on dominance controls.

\begin{table}[!htbp]
\caption{Complete endpoint exchange counts. ``Original'' and ``Exchanged'' count strict outputs; the last four columns partition pairs into candidate-consistent, position-locked, one-strict, and neither-strict outcomes. All position-locked pairs are A/A; none is B/B.}
\label{tab:swap-details}
\centering\footnotesize
\setlength{\tabcolsep}{3pt}
\begin{tabular*}{\textwidth}{@{\extracolsep{\fill}}lrrrrrr@{}}
\toprule
Model & Original & Exchanged & Consistent & Locked & One strict & Neither\\
\midrule
\multicolumn{7}{@{}l}{\emph{Benchmark frontier, 849 pairs}}\\
Qwen3-8B & 188 & 183 & 54 & 0 & 263 & 532\\
Mistral-Small-24B & 119 & 129 & 0 & 0 & 248 & 601\\
Phi-4-14B & 158 & 156 & 52 & 0 & 210 & 587\\
OLMo-Hybrid-7B & 122 & 140 & 0 & 0 & 262 & 587\\
GLM-4-9B & 659 & 668 & 81 & 459 & 247 & 62\\
EXAONE-4.0-32B & 96 & 101 & 72 & 0 & 53 & 724\\
\addlinespace
\multicolumn{7}{@{}l}{\emph{Complete mixed-sign population, 1,345 pairs}}\\
Qwen3-8B & 345 & 351 & 72 & 78 & 396 & 799\\
Mistral-Small-24B & 119 & 129 & 0 & 0 & 248 & 1,097\\
Phi-4-14B & 301 & 301 & 165 & 0 & 272 & 908\\
OLMo-Hybrid-7B & 237 & 244 & 62 & 11 & 335 & 937\\
GLM-4-9B & 1,091 & 1,112 & 176 & 766 & 319 & 84\\
EXAONE-4.0-32B & 140 & 146 & 102 & 0 & 82 & 1,161\\
\bottomrule
\end{tabular*}
\end{table}

Pointwise intervals resample latent tasks 20,000 times, retaining original and exchanged observations together. For the nonzero benchmark-frontier consistency rates, the 95\% intervals are Qwen $[4.57,8.29]$, Phi $[4.36,8.01]$, GLM $[7.57,11.61]$, and EXAONE $[6.52,10.53]$ percent. Complete mixed-sign intervals appear in \Cref{tab:swap-main}. An observed zero gives a degenerate empirical bootstrap distribution, not evidence of a zero population probability. These rates are not causal estimates of the fraction of behavior unaffected by position.

\paragraph{Majority direction and the release set.}
For each canonical vector, majority predicts $\sign(\sum_k x_k)$, with zero supplying no strict direction. The complete mixed-sign population contains 827 strict-majority pairs over 120 distinct canonical profiles and 518 tied-vote pairs over 43 profiles. The benchmark frontier contains 331 of the strict-majority pairs and all 518 tied pairs; the remaining 496 pairs all have a strict majority. This split uses the verified components, not model outputs.

All 577 candidate-consistent model--pair outcomes take the majority direction, with no tied votes. They span 282 distinct pairs and 41 canonical profiles. Within the benchmark frontier, all 259 consistent outcomes, spanning 119 pairs and 13 profiles, also match majority. Pair-weighted and profile-weighted agreement are both 100\% for every nonempty model subset. For profile weighting, we first compute agreement within each signed canonical profile containing consistent outcomes, then weight these profiles equally. Separately, profile-weighted consistency averages the within-profile consistency rate over \emph{all} profiles, including those with no consistent outcomes (\Cref{tab:majority-audit}). Empty consistent subsets have undefined rule agreement.

The tied group has no candidate-consistent strict output from any model, but the single-orientation counts differ. Mistral and EXAONE each return \textsc{underdetermined} in all 1,036 tied-group outputs, with no A, B, \textsc{equal}, or parsing failures. In the strict-majority group, Mistral gives 119 original and 129 exchanged strict verdicts, all labeled A before mapping to canonical identities. EXAONE gives 140 and 146 strict verdicts, respectively; every strict output from both models follows the majority direction. Thus Mistral issues a strict verdict only when displayed A is the majority-favored candidate. Each of the 827 strict-majority pairs presents its majority candidate as A in exactly one orientation. Only 248 of these 827 presentations elicit strict verdicts, so this condition is necessary but not sufficient. Qwen, Phi, OLMo, and GLM also produce strict outputs on tied votes, all favoring displayed A; none of these tied-group outputs survives exchange with candidate identity preserved.

Nine pairs from seven latent tasks are candidate-consistent in all five nonempty model subsets. Their canonical vectors are $(1,1,-1,1,1)$ and its negation: one profile up to sign, outside the benchmark frontier. For the positive orientation, the equal-weight member gives $w_0^\top x=0.2\sum_k x_k=0.60$, whereas $(.1,.1,.6,.1,.1)$ gives $-0.20$. Both weightings are positive and normalized, but the prompt declares neither their relative priority nor majority aggregation. The opposite directions illustrate why selecting the equal-weight member adds a premise. The broader 577-outcome result, rather than this overlap alone, establishes the observed majority pattern.

\begin{table}[!htbp]
\caption{Majority-stratified outputs and profile-weighted consistency. Panel a partitions 1,345 mixed-sign pairs by vote structure; entries count strict outputs or candidate-consistent pairs. Panel b gives the number of profiles with consistent pairs, pair-frequency consistency (\%), and consistency averaged equally across all observed canonical profiles (\%). The complete population has 163 profiles; the benchmark frontier has 85. Conditional on a nonempty consistent subset, majority-direction agreement is 100\% under both weightings.}
\label{tab:majority-audit}
\centering\footnotesize
\setlength{\tabcolsep}{3pt}
\begin{tabular*}{\textwidth}{@{\extracolsep{\fill}}lrrrrrr@{}}
\toprule
\multicolumn{7}{@{}l}{\emph{a\quad Strict outputs by vote structure}}\\
& \multicolumn{3}{c}{Strict majority: 827 pairs} & \multicolumn{3}{c}{Tied majority: 518 pairs}\\
\cmidrule(lr){2-4}\cmidrule(lr){5-7}
Model & Original & Exchanged & Consistent & Original & Exchanged & Consistent\\
\midrule
Qwen3-8B & 277 & 287 & 72 & 68 & 64 & 0\\
Mistral-Small-24B & 119 & 129 & 0 & 0 & 0 & 0\\
Phi-4-14B & 281 & 284 & 165 & 20 & 17 & 0\\
OLMo-Hybrid-7B & 229 & 234 & 62 & 8 & 10 & 0\\
GLM-4-9B & 709 & 723 & 176 & 382 & 389 & 0\\
EXAONE-4.0-32B & 140 & 146 & 102 & 0 & 0 & 0\\
\addlinespace
\multicolumn{7}{@{}l}{\emph{b\quad Candidate consistency: pair-frequency and distinct-profile summaries}}\\
& \multicolumn{3}{c}{Complete mixed-sign population} & \multicolumn{3}{c}{Benchmark frontier}\\
\cmidrule(lr){2-4}\cmidrule(lr){5-7}
Model & Profiles & Pair (\%) & Profile (\%) & Profiles & Pair (\%) & Profile (\%)\\
\midrule
Qwen3-8B & 9 & 5.35 & 5.52 & 5 & 6.36 & 5.88\\
Mistral-Small-24B & 0 & 0.00 & 0.00 & 0 & 0.00 & 0.00\\
Phi-4-14B & 24 & 12.27 & 14.72 & 4 & 6.12 & 4.71\\
OLMo-Hybrid-7B & 8 & 4.61 & 4.91 & 0 & 0.00 & 0.00\\
GLM-4-9B & 25 & 13.09 & 14.87 & 9 & 9.54 & 10.59\\
EXAONE-4.0-32B & 13 & 7.58 & 7.98 & 9 & 8.48 & 10.59\\
\bottomrule
\end{tabular*}
\end{table}

\paragraph{What additive fits establish.}
A strict choice with canonical sign $s$ imposes $s\,w^\top x>0$ on one global nonnegative normalized weight vector. Deferrals impose no constraint; these fits assess only the directions of emitted strict verdicts. We fit original strict choices, both orientations' strict choices, and the candidate-consistent subset separately (\Cref{tab:swap-fit}). Each position-locked pair imposes opposite strict inequalities, forcing at least one exclusion. Exact rational witnesses attain this lower bound in every combined-orientation fit: 78 for Qwen, 11 for OLMo, 766 for GLM, and zero for the others. Mistral illustrates the fit's scope: equal weights fit all 248 strict directions without determining which presentations elicit them. Within each nonempty candidate-consistent subset, zero exclusions follow immediately from the equal-weight member, because every observed direction matches majority. That column is therefore a consequence of the majority result, not independent evidence of inferred internal weights.

\begin{table}[!htbp]
\caption{Endpoint additive fits on all 1,345 mixed-sign pairs. Entries give the certified minimum exclusion count / number of strict observations in a global additive fit. Combined-orientation minima equal the position-locked pair counts in \Cref{tab:swap-details}. Candidate-consistent pairs enter once; equal weights fit all their directions. Dashes denote empty subsets.}
\label{tab:swap-fit}
\centering\footnotesize
\setlength{\tabcolsep}{5pt}
\begin{tabular*}{\textwidth}{@{\extracolsep{\fill}}lrrr@{}}
\toprule
Model & Original & Both orientations & Candidate-consistent\\
\midrule
Qwen3-8B & 41 / 345 & 78 / 696 & 0 / 72\\
Mistral-Small-24B & 0 / 119 & 0 / 248 & --\\
Phi-4-14B & 0 / 301 & 0 / 602 & 0 / 165\\
OLMo-Hybrid-7B & 4 / 237 & 11 / 481 & 0 / 62\\
GLM-4-9B & 379 / 1,091 & 766 / 2,203 & 0 / 176\\
EXAONE-4.0-32B & 0 / 140 & 0 / 286 & 0 / 102\\
\bottomrule
\end{tabular*}
\end{table}

On the 849-pair benchmark frontier, GLM requires 216 exclusions among 659 original strict judgments and 459 among 1,327 combined judgments. Every other model's original and combined strict subsets have zero fit violations. Mistral and OLMo have no candidate-consistent pair in this subset.

\paragraph{Fixed-coordinate and priority-scan comparisons.}
We compare majority with each fixed-coordinate rule $x_j$, $j=1,\ldots,5$ (\Cref{tab:coordinate-audit}). A zero coordinate supplies no direction and counts as a mismatch to an observed strict choice; it is not omitted. No single fixed coordinate matches all candidate-consistent directions for Phi, GLM, or EXAONE in the complete population. Qwen's directions also match slots 4 and 5, and OLMo's match slots 1 and 5. Across all 577 outcomes, slots 1--5 match 301, 387, 403, 358, and 441 directions, respectively; none reproduces the shared majority pattern. Restricting to the benchmark frontier leaves more individual-model rules indistinguishable, as the second panel shows.

We also check fixed scans that choose the first or last nonzero coordinate. Both scans match all OLMo consistent directions; the last-nonzero scan also matches Qwen and EXAONE. Neither scan matches every consistent direction for Phi or GLM. Majority with either scan used only to break vote ties necessarily retains 100\% conditional agreement: no consistent outcome has a tied vote. Thus these tie-breaking variants add no evidence for a particular tie rule. The fixed input order is correctness, completeness, format, concision, and lexical compliance, but the vector prompt omits these names. Slot-based agreement therefore cannot establish a semantic preference such as correctness priority.

\begin{table}[!htbp]
\caption{Fixed-coordinate agreement with candidate-consistent strict directions. Each entry is pair-weighted / profile-weighted agreement (\%), conditional on the model's consistent subset. Both opposite directions and zero coordinates count as mismatches. Dashes denote empty subsets.}
\label{tab:coordinate-audit}
\centering\footnotesize
\setlength{\tabcolsep}{3pt}
\begin{tabular*}{\textwidth}{@{\extracolsep{\fill}}lrrrrr@{}}
\toprule
Model & Slot 1 & Slot 2 & Slot 3 & Slot 4 & Slot 5\\
\midrule
\multicolumn{6}{@{}l}{\emph{Complete mixed-sign population, 1,345 pairs}}\\
Qwen3-8B & 12.50 / 22.22 & 43.06 / 55.56 & 75.00 / 55.56 & 100 / 100 & 100 / 100\\
Mistral-Small-24B & -- & -- & -- & -- & --\\
Phi-4-14B & 63.64 / 75.00 & 56.97 / 70.83 & 71.52 / 58.33 & 66.06 / 58.33 & 83.03 / 75.00\\
OLMo-Hybrid-7B & 100 / 100 & 66.13 / 75.00 & 59.68 / 50.00 & 48.39 / 50.00 & 100 / 100\\
GLM-4-9B & 53.98 / 64.00 & 89.77 / 92.00 & 65.91 / 60.00 & 33.52 / 36.00 & 45.45 / 44.00\\
EXAONE-4.0-32B & 29.41 / 30.77 & 61.76 / 69.23 & 76.47 / 69.23 & 86.27 / 84.62 & 88.24 / 84.62\\
\addlinespace
\multicolumn{6}{@{}l}{\emph{Benchmark frontier, 849 pairs}}\\
Qwen3-8B & 0 / 0 & 24.07 / 20.00 & 100 / 100 & 100 / 100 & 100 / 100\\
Mistral-Small-24B & -- & -- & -- & -- & --\\
Phi-4-14B & 0 / 0 & 25.00 / 25.00 & 100 / 100 & 100 / 100 & 100 / 100\\
OLMo-Hybrid-7B & -- & -- & -- & -- & --\\
GLM-4-9B & 0 / 0 & 77.78 / 77.78 & 100 / 100 & 53.09 / 55.56 & 55.56 / 55.56\\
EXAONE-4.0-32B & 0 / 0 & 75.00 / 77.78 & 79.17 / 77.78 & 80.56 / 77.78 & 83.33 / 77.78\\
\bottomrule
\end{tabular*}
\end{table}

\FloatBarrier

\subsection{Exhaustive profile audit}
\label[appendix]{app:lattice-audit}
The audit in \Cref{sec:configuration} covers 180 mixed-sign profiles, 62 dominance profiles, and one all-equal profile, each in both A/B orientations. The three anonymous encodings use categorical \texttt{A\_ONLY}/\texttt{B\_ONLY}/\texttt{SAME} entries, signed $-1/0/+1$ entries, or a five-row criterion table. All prompts apply the rule in \Cref{eq:pareto-rule} to the same complete evidence, without task or response text.

We reuse the Qwen3-8B and EXAONE-4.0-32B revisions in \Cref{tab:models} and add Qwen3.8-27B at revision \seqsplit{1d4bf0f2ff6012fd82039f2fa52739d0dd7c60c0}. Each prompt uses seeds 20260919, 20260920, and 20260921 in both modes, with sampling settings and the 2,048-token cap matched within each model. Temperatures for Qwen3-8B, EXAONE, and Qwen3.8 are $0.6,0.6,1.0$, with batch sizes 2, 1, and 1, respectively. All runs use bfloat16 and top-$p=0.95$; top-$k$ is 20 for both Qwen checkpoints and disabled for EXAONE. Reasoning-enabled (ON) and reasoning-disabled (OFF) configurations use the corresponding official chat templates. Qwen3.8 ON additionally sets \texttt{reasoning\_effort=medium}.

Comparison with the primary greedy audit in \Cref{app:tradeoff-decoding} also changes sampling settings and the output cap.
Equal token caps do not imply equal realized generation lengths or compute, and this experiment does not separate the contribution of the reasoning trace from other mode-dependent behavior.

Completion lengths include reasoning but exclude input and padding tokens. All 13,122 ON outputs contain nonempty reasoning.
Each model has 4,374 outputs per mode ($243$ profiles $\times$ two orientations $\times$ three encodings $\times$ three seeds).
\Cref{tab:reasoning-lengths} reports the observed maxima.
All 26,244 outputs terminate with an end-of-sequence marker before reaching the matched 2,048-token cap.
These maximum lengths are descriptive, not estimates of mean token use, paired overhead, latency, energy, or end-to-end savings.
Token counts use each model's tokenizer, so comparisons should be made within each model.

\begin{table}[!htbp]
\caption{Observed maximum completion length by mode in the exhaustive profile audit. Each entry is a maximum over 4,374 outputs, not a mean or a matched-example overhead. Lengths include reasoning where present and exclude input and padding tokens. The maximum allowance is 2,048 tokens in both modes.}
\label{tab:reasoning-lengths}
\centering\small
\begin{tabular}{lrr}
\toprule
Model & OFF maximum & ON maximum\\
\midrule
Qwen3-8B & 16 & 1,419\\
EXAONE-4.0-32B & 17 & 1,511\\
Qwen3.8-27B & 16 & 293\\
\bottomrule
\end{tabular}
\end{table}

\begin{table}[!htbp]
\caption{Exhaustive profile audit by model, mode, and encoding. Mixed strict gives unsupported A/B verdicts over mixed-sign orientation decisions, with percentages in parentheses. Consistent counts mixed-sign profile--encoding--seed events in which both orientations select the same canonical candidate. Dominance and equality give correct/total decisions.}
\label{tab:lattice-audit}
\centering\footnotesize
\setlength{\tabcolsep}{3pt}
\begin{tabular*}{\textwidth}{@{\extracolsep{\fill}}lllrrrr@{}}
\toprule
Model & Mode & Encoding & Mixed strict & Consistent & Dominance & Equality\\
\midrule
Qwen3-8B & OFF & Categorical & 853/1,080 (78.98) & 80 & 334/372 & 6/6\\
 & & Signed numeric & 306/1,080 (28.33) & 16 & 332/372 & 6/6\\
 & & Criterion table & 84/1,080 (7.78) & 3 & 365/372 & 6/6\\
 & ON & Categorical & 1/1,080 (0.09) & 0 & 372/372 & 6/6\\
 & & Signed numeric & 0/1,080 (0) & 0 & 372/372 & 6/6\\
 & & Criterion table & 0/1,080 (0) & 0 & 372/372 & 6/6\\
\addlinespace
EXAONE-4.0-32B & OFF & Categorical & 291/1,080 (26.94) & 118 & 366/372 & 6/6\\
 & & Signed numeric & 407/1,080 (37.69) & 146 & 366/372 & 6/6\\
 & & Criterion table & 34/1,080 (3.15) & 6 & 363/372 & 6/6\\
 & ON & Categorical & 0/1,080 (0) & 0 & 372/372 & 6/6\\
 & & Signed numeric & 0/1,080 (0) & 0 & 372/372 & 6/6\\
 & & Criterion table & 1/1,080 (0.09) & 0 & 372/372 & 6/6\\
\addlinespace
Qwen3.8-27B & OFF & Categorical & 63/1,080 (5.83) & 7 & 372/372 & 6/6\\
 & & Signed numeric & 37/1,080 (3.43) & 0 & 372/372 & 6/6\\
 & & Criterion table & 0/1,080 (0) & 0 & 372/372 & 6/6\\
 & ON & Categorical & 0/1,080 (0) & 0 & 372/372 & 6/6\\
 & & Signed numeric & 0/1,080 (0) & 0 & 372/372 & 6/6\\
 & & Criterion table & 0/1,080 (0) & 0 & 372/372 & 6/6\\
\bottomrule
\end{tabular*}
\end{table}

The OFF candidate-consistent events in \Cref{tab:lattice-audit} cover 36, 69, and four distinct canonical profiles for Qwen3-8B, EXAONE, and Qwen3.8, respectively. EXAONE OFF selects the same canonical candidate in both orientations and all three encodings for four profile--seed combinations spanning three distinct profiles. Signed-numeric violations rule out counting only literal \texttt{A\_ONLY}/\texttt{B\_ONLY} evidence tokens, but not counting signs or another shallow aggregation heuristic. Some OFF cells also fail dominance controls, so their mixed-sign rates do not isolate conflict-specific errors. Counts describe this finite grid; repeated orientations, encodings, and seeds are not independent profile observations.

All outputs are parseable except two EXAONE OFF dominance responses returning \texttt{B\_ONLY}; these invalid verdicts remain in the denominator as errors. The two ON violations each occur in only one A/B orientation. Qwen3-8B overlooks a \texttt{B\_ONLY} entry, whereas EXAONE acknowledges a four-to-one conflict but outputs its majority direction.

\subsection{Executable identification alongside either inference mode}
\label[appendix]{app:reasoning-integration}
This interface implements the existing decision rule; it is not a new learned judge or an additional model experiment.
The input is a complete, validated component vector and a versioned declared specification.
For TradeoffBench, the four-way implementation of \Cref{eq:pareto-rule} is:
\begin{quote}
\small
\begin{verbatim}
if every coordinate equals zero: return EQUAL
if some coordinate is positive and none is negative: return A
if some coordinate is negative and none is positive: return B
otherwise: return UNDERDETERMINED
\end{verbatim}
\end{quote}
On the $3^5$ ternary domain, this rule gives 31 A profiles, 31 B profiles, one EQUAL profile, and 180 UNDERDETERMINED profiles.
These counts follow from the stated rule, not from new LLM inference.
Both cases represented by zero in the mathematical rule withhold a strict direction, while the interface distinguishes equal components from unresolved conflict.

For general polyhedral $\W$, compute the two utility bounds in \Cref{eq:robust-interval} and apply \Cref{eq:core}. HUMAINE's symmetric specification also has the closed form in \Cref{eq:humaine-closed-form}.
A strict release requires a verified positive lower bound or negative upper bound.
Incomplete inputs, an infeasible specification, or an unresolved numerical check require a separate input or solver-status flag, not a strict verdict or an assertion of genuine non-identification.

A reasoning-enabled judge may assess components or help formulate a proposed refinement.
Record the component source, coding, specification version, interval or rule result, and any added premise.
Make any refinement explicit before recomputing identification. Agreement with a target remains a separate policy-level audit.
This interface does not require switching reasoning off.
If the reasoning-enabled judge already executes the same rule correctly on the same inputs, the interface leaves the verdict unchanged. It adds an explicit support record and a defined route for further resolution.

The strict Pareto implementation scans $d$ coordinates in $O(d)$ operations and needs no further model generation after the structured inputs are available.
This statement excludes component acquisition, preference elicitation, general LP solution, risk calibration, and review.
It establishes the scope of the aggregation computation, not an end-to-end efficiency advantage or a measured improvement in target agreement with reasoning enabled.

\FloatBarrier

\section{MultiPref: Component Substitution and Membership}
\label[appendix]{app:multipref}

MultiPref contains 10,461 comparisons, 41,844 annotations, and four evaluators per comparison. Its three component criteria are helpfulness, truthfulness, and harmlessness. Component preferences are encoded as $1,0.5,0,-0.5,-1$, with zero treated as coordinate equality. Overall preferences are obtained by collapsing response-strength labels to $-1,0,+1$. A global-time cutoff and prompt exclusion yield disjoint development and evaluation partitions. Eighty evaluators meet the minimum of 50 development and 10 evaluation annotations. The main individual analysis contains 4,847 target-evaluator decisions over 3,561 comparisons and 2,939 prompts, including 3,515 strict overall targets.

Both conditions apply the strict Pareto rule in \Cref{eq:pareto-rule} to three components. The own-component condition uses the target evaluator's components. The peer-mean condition averages the other three evaluators' components coordinatewise, preserving an average of zero as coordinate equality. The overall target is unchanged. The earlier partition establishes evaluator eligibility; the main substitution contrast uses the evaluation partition. Coverage uses 4,847 eligible decisions, and strict-target coverage uses their 3,515 strict targets. Strict-direction risk is computed among the strict targets released by each core. Inference jointly resamples evaluator and prompt clusters. Component source and averaging change together in this original contrast; \Cref{app:single-peer} adds a single-peer condition to separate substitution from averaging.

\subsection{Separating component substitution from peer averaging}
\label[appendix]{app:single-peer}
The single-peer ablation uses the original eligible MultiPref evaluation partition from \Cref{app:multipref}. A stable hash of the seed, comparison, target evaluator, and peer evaluator orders the three peers without using their outcomes. The first peer supplies the primary single-peer condition; the other two ranks are descriptive sensitivity checks. The own, single-peer, and peer-mean rules share 2,175 released strict targets. Holding these targets fixed prevents released-set changes from explaining the primary risk contrast.

\begin{table}[!htbp]
\caption{Single-peer ablation on the same 2,175 strict targets, with the intended evaluator's overall preference and strict Pareto rule fixed. The original own/peer-mean result on 2,707 targets remains a separate comparison.}
\label{tab:single-peer}
\centering\footnotesize
\begin{tabular*}{\textwidth}{@{\extracolsep{\fill}}lrr@{}}
\toprule
Component source & Directional errors & Strict-direction risk (\%)\\
\midrule
Own & 12 & 0.55\\
Single peer & 651 & 29.93\\
Peer mean & 590 & 27.13\\
\bottomrule
\end{tabular*}
\end{table}

We use 5,000 paired crossed target-evaluator/prompt bootstrap replicates. Three-contrast Bonferroni-adjusted percentile intervals give a single-peer-minus-own difference of $29.38$ points, $[24.75,34.10]$, and a peer-mean-minus-own difference of $26.57$ points, $[22.18,31.04]$. The peer-mean-minus-single-peer difference is $-2.80$ points, $[-5.59,0.18]$. Thus averaging is not required for the large substitution gap; these data do not establish an averaging effect under the three-contrast adjustment. The intervals are approximate and describe agreement with the intended rater.

\subsection{Complete release and risk accounting}
\begin{figure}[!htbp]
\centering
\includegraphics[width=\textwidth]{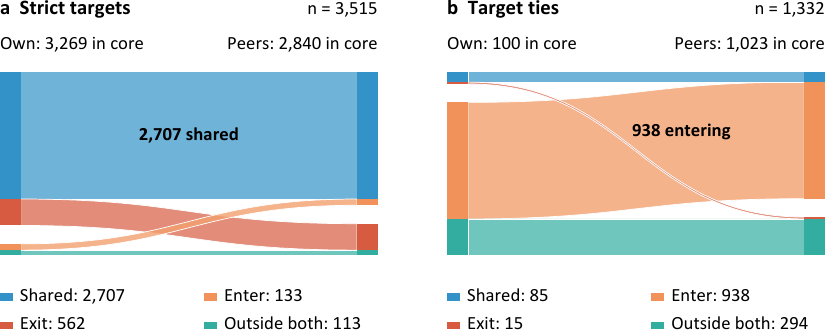}
\caption{The net increase in core membership after substitution is concentrated among target ties. ``Peers'' denotes peer-mean components. Ribbons connect core membership under own and peer-mean components, separately for strict targets (a) and target ties (b). Upper nodes are in the core; lower nodes are outside it. Widths are proportional within each panel, and count keys retain all four membership strata. Target labels are fixed. Among strict targets, the own-only core has seven errors and the peer-only core has 67; shared-core errors are reported in \Cref{tab:transport-all}.}
\label{fig:core-membership}
\end{figure}

\Cref{fig:core-membership} shows the composition changes underlying the aggregate rows in \Cref{tab:transport-all}. Substitution adds 1,071 core members and removes 577, but 938 of the additions (87.6\%) have target ties. Released strict targets therefore fall from 3,269 to 2,840 despite the increase in all-record coverage. The shared-core comparison holds the 2,707 common strict targets fixed, separating its risk contrast from this membership change.

\begin{table}[!htbp]
\caption{Strict-direction risk under the strict Pareto rule using the target evaluator's own components or the other three evaluators' mean. Overall target labels and the aggregation rule are fixed. Coverage uses 4,847 eligible evaluator decisions; strict-direction risk uses released strict targets in each row. The shared-core rows include 85 target ties; the main paired-risk contrast uses their 2,707 strict targets (\Cref{fig:transport-main}).}
\label{tab:transport-all}
\centering\footnotesize
\setlength{\tabcolsep}{4pt}
\begin{tabular*}{\textwidth}{@{\extracolsep{\fill}}llrrrr@{}}
\toprule
Support & Components & All releases & Coverage (\%) & Strict rel./err. & Risk (\%)\\
\midrule
All eligible & Own & 3,369 & 69.51 & 3,269 / 23 & 0.704\\
All eligible & Peer mean & 3,863 & 79.70 & 2,840 / 847 & 29.824\\
\addlinespace
Shared core & Own & 2,792 & -- & 2,707 / 16 & 0.591\\
Shared core & Peer mean & 2,792 & -- & 2,707 / 780 & 28.814\\
\bottomrule
\end{tabular*}
\end{table}

\FloatBarrier
\subsection{Paired directional changes on shared targets}
On the same 2,707 strict targets, 1,919 decisions agree with the target under both component conditions and eight oppose it under both. The other 780 records consist of 772 changes from agreement to opposition and eight from opposition to agreement. Thus own components have $8+8=16$ errors, while peer-mean components have $772+8=780$ errors. The difference of 764 errors on this fixed denominator corresponds to 28.223 percentage points in \Cref{fig:transport-main}. Joint evaluator--prompt resampling retains the within-record pairing (\Cref{sec:artifact}).

\subsection{Consistency across evaluators and time partitions}
\label[appendix]{app:multipref-consistency}
To assess variation across evaluators, we use all eligible evaluation records rather than restricting to the shared core. For each evaluator, residual error rate is the number of erroneous strict releases divided by that evaluator's total number of strict targets. Own components yield a lower residual error rate for 73 of the 80 evaluators, equal error rate for seven, and higher error rate for none. This comparison uses a common denominator within each evaluator, while allowing core membership to differ between component conditions.

Applying the same coding and strict Pareto rule to the earlier partition provides a within-dataset temporal consistency check. Among 7,337 decisions from the same 80 evaluators, 5,394 have strict targets. Own components yield 37 errors in 5,111 strict releases (0.724\%); peer-mean components yield 1,307 errors in 4,410 (29.637\%). The corresponding evaluation risks are 0.704\% and 29.824\% (\Cref{tab:transport-all}). These rates condition on each core's released strict targets, unlike the evaluator-level residual-error denominator above.

\FloatBarrier

\section{HUMAINE: Geometry, Qualification, and Review}
\label[appendix]{sec:humaine-details}
This appendix distinguishes three analyses: the releases of each fixed rule, qualification and selection over candidate policies, and prioritization of a fixed review workload. The geometry and numerical checks support \Cref{sec:result-humaine}. Contract sensitivities explain conditional selection, and ranking comparisons assess whether the structural frontier also provides an error ordering.

\subsection{Core geometry and declared values}
The lower bounds and normalization in \Cref{eq:scalar} imply $w_k\leq1-3(0.20)=0.40$. Writing $w=(\mathbf1+v)/5$ with $v\geq0$ and $\mathbf1^\top v=1$ expresses $\Wp$ as the convex hull of four criterion-emphasis profiles. Each vertex assigns weight .40 to one criterion and weight .20 to each of the other three. The minimum and maximum over these vertices are
\begin{equation}
 \underline u_{\Wp}(x)=0.2\left(\sum_k x_k+\min_k x_k\right),
 \qquad
 \overline u_{\Wp}(x)=0.2\left(\sum_k x_k+\max_k x_k\right).
 \label{eq:humaine-closed-form}
\end{equation}
Under the ternary coding $x_k\in\{-1,0,+1\}$, the positive core extends strict Pareto exactly by profiles with three components favoring A and one favoring B. The negative case is symmetric. Every added profile has equal-weight margin 0.50, giving the exact hybrid representation in \Cref{eq:humaine-hybrid}.

\rev{Recall the asymmetric safety-protective set $\Ws$ defined in \Cref{eq:humaine-wsafe} (\Cref{sec:result-humaine}).} For a three-versus-one profile, let $s=\sign(w_0^\top x)$ and let $j$ be its sole opposing criterion. Then $w^\top(sx)=1-2w_j$, giving the minimum gaps in \Cref{tab:value-boundary}. Among the retained three-versus-one records, the opposing criterion is Core Task Performance \& Reasoning in 94, Communication Style \& Presentation in 208, and Interaction Fluidity \& Adaptiveness in 216. The safety-protective core releases 24,250 judgments: the 23,732 Pareto releases and these 518 additions, with no directional conflict.

\begin{table}[!htbp]
\caption{An asymmetric weight set distinguishes conflicts with equal scalar margins. These are the three-versus-one additions to Pareto in HUMAINE. Let $s=\sign(w_0^\top x)$ orient each profile toward its majority. For sole opposing criterion $j$, the minimum majority-aligned gap is $\underline u_{\Ws}(sx)=1-2\max_{w\in\Ws}w_j$.}
\label{tab:value-boundary}
\centering\footnotesize
\setlength{\tabcolsep}{4pt}
\begin{tabular*}{\textwidth}{@{\extracolsep{\fill}}lrrcrc@{}}
\toprule
& & & $\Wp$ & \multicolumn{2}{c}{$\Ws$}\\
\cmidrule(lr){5-6}
Opposing criterion & Records & $|w_0^\top x|$ & Decision & Min. gap & Decision\\
\midrule
Non-safety & 518 & 0.50 & Release & $+0.20$ & Release\\
Trust, Ethics \& Safety & 238 & 0.50 & Release & $-0.10$ & Defer\\
\bottomrule
\end{tabular*}
\end{table}

\subsection{Exact weight-set sensitivity}
\label[appendix]{app:humaine-value-path}
The symmetric path varies the minimum emphasis shared by all four criteria:
\begin{equation}
 \W_{\ell}=\{w\in\R_+^4:\mathbf1^\top w=1,\;\ell\leq w_k\leq1-3\ell\},
 \qquad 0\leq\ell\leq0.25,
 \label{eq:humaine-path}
\end{equation}
Writing $w=\ell\mathbf1+(1-4\ell)v$ with $v\geq0$ and $\mathbf1^\top v=1$ yields
\begin{equation}
\begin{aligned}
\underline u_{\W_\ell}(x)&=\ell\sum_kx_k+(1-4\ell)\min_kx_k,\\
\overline u_{\W_\ell}(x)&=\ell\sum_kx_k+(1-4\ell)\max_kx_k.
\end{aligned}
\label{eq:humaine-path-utilities}
\end{equation}
For each of the 81 ternary profiles, these bounds are affine in $\ell$. Their zero crossings give the complete partition in \Cref{tab:humaine-value-path}. Decisions cannot change within an open interval between crossings. In particular, the minimum utilities for $(1,1,1,-1)$ and $(1,1,0,-1)$ are $6\ell-1$ and $5\ell-1$, respectively. Allowing the former compensation while withholding the latter therefore corresponds exactly to $1/6<\ell\leq1/5$. At $\ell=1/6$, the three-versus-one minimum is zero; at $\ell=1/5$, the two-versus-one minimum is zero. Neither zero supports a strict release. Thus $\Wp=\W_{1/5}$ is the upper endpoint of this policy interval. Increasing $\ell$ beyond it changes the rule.

\begin{table}[!htbp]
\caption{The complete symmetric value path has four decision states. The partition is exact on all $3^4=81$ ternary profiles, including equalities and both directions. Counts describe profiles, not independent observations. Scalar rules are defined in \Cref{eq:scalar}.}
\label{tab:humaine-value-path}
\centering\footnotesize
\setlength{\tabcolsep}{8pt}
\begin{tabular}{@{}llr@{}}
\toprule
Parameter range & Equivalent release rule & Released profiles / 81\\
\midrule
$\ell=0$ & E1.00 & 2\\
$0<\ell\leq1/6$ & Strict Pareto & 30\\
$1/6<\ell\leq1/5$ & Symmetric partial & 38\\
$1/5<\ell\leq1/4$ & E.25 & 62\\
\bottomrule
\end{tabular}
\end{table}

All four states belong to the existing seven-candidate simultaneous family. Under Bonferroni calibration at the 2.5\% pooled strict-direction tolerance, the symmetric partial core has the highest coverage among passing states on this path (\Cref{tab:candidate-audit}). The continuous characterization therefore requires no additional candidate rules or multiplicity correction.

Exact rational arithmetic and independent vertex enumeration agree on all 81 profiles. Replaying both HUMAINE releases reproduces the seven candidates' counts, calibration bounds, and policy selections. Throughout the partial interval, release directions and the review queue in \Cref{app:review-allocation} are identical. These checks establish decision equivalence on the existing data, not independent replications.

The numerical audit uses 39 settings, including the twelve grid settings, exact transition points, interval interiors, and perturbations near the transitions. All twelve grid settings give the same directions under LP and exact arithmetic. Additional near-boundary probes expose conservative deferrals from the $10^{-10}$ release tolerance and a maximum LP utility error of $4\times10^{-8}$, with no unsupported strict releases. The partition in \Cref{tab:humaine-value-path} uses exact signs; the numerical checks preserve the original solver and tolerance.

\subsection{Global risk contracts and tie-aware calibration}
\label[appendix]{app:calibration-libraries}
Each non-nested candidate library contains E.25, E.50, and E.75 plus one alternative: E1.00 for scalar-only, strict Pareto for Pareto-first, the symmetric partial core for partial-value, or the safety-protective core for safety-protective. The supplementary nested comparison includes Pareto in the partial-value and safety-protective libraries to examine the effect of different fallback candidates. Both designs use the same seven distinct candidates and shared calibration tests. Nesting ensures nondecreasing maximum qualifying calibration coverage under fixed candidate tests; held-out coverage and risk remain empirical comparisons. Selection compares complete policies, and qualification concerns each policy's specified target-risk contract.

Pooled Bonferroni calibration uses a per-candidate confidence level of $1-0.05/7$. Within each library, selection maximizes all-record calibration coverage among candidates whose earlier-release upper bound passes the tolerance. \Cref{tab:candidate-audit} reports both this selection objective and the strict-target counts used for the risk bound. Evaluation metrics use the later release.

\begin{table}[!htbp]
\caption{Complete HUMAINE candidate audit for pooled strict-direction risk. All seven candidates enter one Bonferroni family. All-record coverage uses 21,044 calibration or 30,425 evaluation records. ``Rel./err.'' gives released strict targets and opposite-direction errors. Coverage, UCB, and risk are percentages.}
\label{tab:candidate-audit}
\centering
\footnotesize
\begin{tabular}{@{}lrrrrrr@{}}
\toprule
& \multicolumn{3}{c}{Calibration} & \multicolumn{3}{c}{Evaluation}\\
\cmidrule(lr){2-4}\cmidrule(lr){5-7}
Candidate & All cov. & Rel./err. & UCB & All cov. & Rel./err. & Risk\\
\midrule
E.25 & 88.738 & 17,621 / 404 & 2.584 & 87.790 & 25,162 / 487 & 1.935 \\
E.50 & 66.922 & 13,734 / 74 & 0.712 & 65.305 & 19,373 / 87 & 0.449 \\
E.75 & 39.009 & 8,138 / 0 & 0.061 & 37.377 & 11,255 / 0 & 0.000 \\
E1.00 & 18.214 & 3,820 / 0 & 0.129 & 17.302 & 5,237 / 0 & 0.000 \\
Strict Pareto & 77.827 & 15,543 / 173 & 1.337 & 78.002 & 22,437 / 196 & 0.874 \\
Symmetric partial & 80.783 & 16,140 / 199 & 1.462 & 80.486 & 23,160 / 224 & 0.967 \\
Safety-protective & 79.895 & 15,961 / 192 & 1.431 & 79.704 & 22,935 / 210 & 0.916 \\
\bottomrule
\end{tabular}
\end{table}

\paragraph{Where the added releases occur.}
\Cref{tab:partial-layer} \rev{in \Cref{sec:humaine-release}} decomposes the symmetric partial release into the Pareto core and its three-versus-one additions. The added layer contains 723 strict targets and 28 directional errors, an observed risk of 3.873\%. The complete policy has 224 errors among 23,160 released strict targets, or 0.967\%. The pooled contract qualifies this complete policy; it does not certify the added layer at the 2.5\% tolerance.

\label[appendix]{app:participant-linkage}
Layer bounds use one-sided Clopper--Pearson limits at $1-.05/3$. Bonferroni does not require independence between layers; each binomial bound assumes independent comparisons. The releases provide conversation keys but no participant linkage across conversations; participants can return across batches \citep{petrova2026humaine}. The five metric rows per conversation are already combined into one comparison. Participant-cluster correction is therefore unavailable; \Cref{app:calibration-sensitivity} reports hypothetical information reductions.

For the shared worst-group contract in \Cref{tab:global-contracts}, correction covers the $6\times7$ group--candidate family. One global candidate is selected only if it passes every group-specific bound.

\begin{table}[!htbp]
\caption{HUMAINE selection under different risk contracts with Bonferroni correction. Entries give the calibration-selected rule / evaluation all-record coverage (\%). Partial and Safety denote the symmetric and safety-protective cores. Every contract deploys one rule globally; shared worst-group selection additionally requires that rule to pass all six group bounds.}
\label{tab:global-contracts}
\centering
\footnotesize
\setlength{\tabcolsep}{2pt}
\begin{tabular*}{\textwidth}{@{\extracolsep{\fill}}lrcccc@{}}
\toprule
Contract / loss & Tolerance & Scalar-only & Pareto-first & Partial-value & Safety-protective \\
\midrule
Pooled / directional & 2.5\% & E.50 / 65.305 & Pareto / 78.002 & Partial / 80.486 & Safety / 79.704 \\
Shared worst-group / directional & 2.5\% & E.50 / 65.305 & E.50 / 65.305 & E.50 / 65.305 & E.50 / 65.305 \\
\midrule
Pooled / tie-aware & 2.5\% & E.75 / 37.377 & E.75 / 37.377 & E.75 / 37.377 & E.75 / 37.377 \\
& 5.0\% & E.50 / 65.305 & E.50 / 65.305 & E.50 / 65.305 & E.50 / 65.305 \\
& 7.5\% & E.50 / 65.305 & Pareto / 78.002 & Partial / 80.486 & Safety / 79.704 \\
& 10.0\% & E.25 / 87.790 & E.25 / 87.790 & E.25 / 87.790 & E.25 / 87.790 \\
\bottomrule
\end{tabular*}
\end{table}

\subsection{Multiplicity correction and assumed effective sample size}
\label[appendix]{app:calibration-sensitivity}
We compare Bonferroni and Holm correction at eight assumed effective sample fractions: 1, .75, .50, .25, .10, .075, .05, and .025. Candidate rules, libraries, and selection objectives remain fixed. Both losses use tolerance .025 and family error level .05. Each loss, contract, and assumed fraction defines a separate comparison: seven tests for pooled qualification and 42 for shared worst-group qualification.

For a candidate with $k$ calibration errors among $n$ released targets eligible for the loss, we test the null that its risk is at least .025 using
\begin{equation}
p(k,n)=\Pr\{\operatorname{Binomial}(n,.025)\leq k\}.
\label{eq:calibration-test}
\end{equation}
Bonferroni compares each value with $.05/m$, where $m$ is the family size. Holm sorts them as $p_{(1)}\leq\cdots\leq p_{(m)}$ and rejects sequentially while $p_{(i)}\leq.05/(m-i+1)$, stopping at the first failure \citep{holm1979simple}. A shared worst-group candidate must pass all six group tests. Selection then maximizes calibration coverage within the unchanged library. If no candidate passes, the policy defers on every record. Holm eligibility is a family-adjusted test decision, not a candidate-specific Clopper--Pearson bound computed from its observed rank.

At nominal sample size, E.25 has $p=.03967893$ and is last in the seven-rule pooled directional family. It passes Holm after the other six rejections, so every library selects E.25. The Bonferroni bound instead excludes it (\Cref{tab:candidate-audit}). This changes selection without changing the data, risk tolerance, or family error level.

Participant identifiers are unavailable, so the effective-sample analysis is hypothetical. For a nonempty calibration sample, set $n_{\mathrm{eff}}=\max\{1,\lfloor qn\rfloor\}$ and $k_{\mathrm{eff}}=\lceil(k/n)n_{\mathrm{eff}}\rceil$; an empty sample cannot qualify. We recompute tests from these effective counts, leaving evaluation records unchanged. \Cref{tab:correction-ess} reports the pooled directional comparison; \Cref{tab:other-contract-sensitivity} gives the remaining contracts. The two corrections differ at $q=1$, .05, and .025.

\begin{table}[!htbp]
\caption{Correction and hypothetical effective information alter pooled directional selection at 2.5\% tolerance. Coverage (\%) refers to the non-nested libraries. $\Delta_{\mathrm{non}}$ and $\Delta_{\mathrm{nest}}$ give partial-value minus Pareto-first for the non-nested and nested libraries, respectively, in percentage points from unrounded coverage. Each $q$ is a separate assumed information level.}
\label{tab:correction-ess}
\centering\footnotesize
\setlength{\tabcolsep}{4pt}
\begin{tabular*}{\textwidth}{@{\extracolsep{\fill}}rrrrrrrrr@{}}
\toprule
& \multicolumn{4}{c}{Bonferroni} & \multicolumn{4}{c}{Holm}\\
\cmidrule(lr){2-5}\cmidrule(lr){6-9}
$q$ & Pareto-first & Partial-value & $\Delta_{\mathrm{non}}$ & $\Delta_{\mathrm{nest}}$ & Pareto-first & Partial-value & $\Delta_{\mathrm{non}}$ & $\Delta_{\mathrm{nest}}$\\
\midrule
1.000 & 78.002 & 80.486 & $+2.485$ & $+2.485$ & 87.790 & 87.790 & 0 & 0\\
0.750 & 78.002 & 80.486 & $+2.485$ & $+2.485$ & 78.002 & 80.486 & $+2.485$ & $+2.485$\\
0.500 & 78.002 & 80.486 & $+2.485$ & $+2.485$ & 78.002 & 80.486 & $+2.485$ & $+2.485$\\
0.250 & 78.002 & 80.486 & $+2.485$ & $+2.485$ & 78.002 & 80.486 & $+2.485$ & $+2.485$\\
0.100 & 78.002 & 80.486 & $+2.485$ & $+2.485$ & 78.002 & 80.486 & $+2.485$ & $+2.485$\\
0.075 & 78.002 & 80.486 & $+2.485$ & $+2.485$ & 78.002 & 80.486 & $+2.485$ & $+2.485$\\
0.050 & 78.002 & 65.305 & $-12.697$ & 0 & 78.002 & 80.486 & $+2.485$ & $+2.485$\\
0.025 & 37.377 & 37.377 & 0 & 0 & 65.305 & 65.305 & 0 & 0\\
\bottomrule
\end{tabular*}
\end{table}

For Bonferroni at tested $q\geq.075$, scalar-only selects E.50 and safety-protective selects its core, with 65.305\% and 79.704\% coverage. At .05, the non-nested partial-value and safety-protective libraries fall back to E.50 while Pareto still passes; the supplementary nested libraries select Pareto, with 78.002\% coverage. At .025, every library selects E.75. Under Holm, the nominal Bonferroni selections hold at tested $.05\leq q\leq.75$; at .025, all select E.50. All other selections on the tested grid agree across the two designs.

\begin{table}[!htbp]
\caption{Other 2.5\% contracts under hypothetical effective sample sizes. Both corrections give the same selection in every row, and all four libraries select the same rule. Coverage uses all 30,425 evaluation records; defer-all has zero coverage and undefined conditional risk. Ranges refer only to the tested grid.}
\label{tab:other-contract-sensitivity}
\centering\footnotesize
\setlength{\tabcolsep}{5pt}
\begin{tabular*}{\textwidth}{@{\extracolsep{\fill}}lllr@{}}
\toprule
Contract / loss & Tested $q$ & Selected rule & Coverage (\%)\\
\midrule
Shared worst-group / directional & $1,.75$ & E.50 & 65.305\\
 & $.50$ & E.75 & 37.377\\
 & $.025$--$.25$ & Defer all & 0\\
Pooled / tie-aware & $.075$--$1$ & E.75 & 37.377\\
 & $.05,.025$ & Defer all & 0\\
Shared worst-group / tie-aware & All eight settings & Defer all & 0\\
\bottomrule
\end{tabular*}
\end{table}

The break-even calculation in \Cref{tab:ess-sensitivity} resolves the primary Bonferroni selection more finely than this grid. For each nominally passing candidate, $q^*$ is the smallest effective fraction at which it passes for that and all larger effective counts. All four primary selections remain unchanged for $q\geq0.0612$. Integer rounding is conservative at each evaluated count, but this stress test is not a formal dependence guarantee.

\begin{table}[!htbp]
\caption{Break-even effective sample size for the 2.5\% pooled strict-direction bound. The implied design effect is $1/q^*$. A dash denotes a candidate that fails at the nominal sample size.}
\label{tab:ess-sensitivity}
\centering
\footnotesize
\begin{tabular}{lrrr}
\toprule
Candidate & Nominal UCB & Stable $q^*$ & Implied design effect \\
\midrule
E.25 & 2.584\% & -- & -- \\
E.50 & 0.712\% & 0.0304 & 32.94 \\
E.75 & 0.061\% & 0.0241 & 41.52 \\
E1.00 & 0.129\% & 0.0513 & 19.49 \\
Strict Pareto & 1.337\% & 0.0531 & 18.84 \\
Symmetric partial & 1.462\% & 0.0612 & 16.35 \\
Safety-protective & 1.431\% & 0.0585 & 17.09 \\
\bottomrule
\end{tabular}
\end{table}

\subsection{Fixed-budget review allocation}
\label[appendix]{app:review-allocation}
The review workload contains the nonzero outputs of $h_0=s_{.25,w_0}$: 18,674 earlier-release judgments for fitting the risk rankers and 26,710 later-release judgments for evaluation. The remaining 3,715 later records have zero scalar outputs and are excluded by their components, not their target labels. The evaluation workload includes 1,548 target ties. Since the partial core's 24,488 judgments agree with $h_0$, the remaining $26{,}710-24{,}488=2{,}222$ determine the reference review budget using component evidence alone. Additional fractions of 1\%, 2\%, 5\%, 10\%, 15\%, and 20\% are rounded down to integer counts. The seven budgets are $b\in\{267,534,1335,2222,2671,4006,5342\}$ for every priority.

For each initial strict judgment $i$, define directional and tie-aware error indicators by
\begin{equation}
e_{i,\mathrm{dir}}=\mathbf1\{Y_i\neq0,\ h_0(x_i)\neq Y_i\},
\qquad
e_{i,\mathrm{tie}}=\mathbf1\{h_0(x_i)\neq Y_i\}.
\label{eq:review-errors}
\end{equation}
There are 487 directional and 2,035 tie-aware errors. The partial queue contains the $487-224=263$ directional errors excluded by partial release in \Cref{tab:headline-main}. A review priority sorts judgments by decreasing score. Scores tied at the budget boundary are sampled uniformly without replacement. Let $q_i$ be the resulting inclusion probability, so $\sum_i q_i=b$. For loss $j\in\{\mathrm{dir},\mathrm{tie}\}$, expected error capture is
\begin{equation}
\operatorname{Capture}_j(q)=\frac{\sum_i q_i e_{i,j}}{\sum_i e_{i,j}}.
\label{eq:review-capture}
\end{equation}
We use the exact expectation over tied priorities, rather than one random tie-breaking realization. Uniform review uses a constant score, margin priority uses $-|w_0^\top x|$, and Pareto and partial priorities use $\mathbf1\{r_{\mathrm P}(x)=0\}$ and $\mathbf1\{r_{\Spec_{\mathrm p}}(x)=0\}$, respectively. Above frontier capacity, each binary priority fills uniformly from its core. The binary partial-frontier priority is identical to the deferral indicator of the Pareto-plus-margin release implementation in \Cref{eq:humaine-hybrid}, on all 81 ternary profiles and in every bootstrap replicate. This identity is distinct from the within-layer ordering used by Pareto then margin.

Pareto then margin orders lexicographically by decreasing $\mathbf1\{r_{\mathrm P}(x)=0\}$ and then increasing $|w_0^\top x|$. Within the E.25 workload, all partial-frontier profiles are permutations or sign reversals of $(1,1,-1,0)$. Their margin is .25, whereas the other mixed-sign profiles have margin .50. Thus Pareto then margin prioritizes exactly the partial frontier before any other eligible profile. The two priorities have identical inclusion probabilities at every budget no greater than the observed frontier capacity. The linear capture path below this capacity is a consequence of uniform sampling from a fixed region, not a sequence of independent validations.

Each learned ranker predicts an error indicator from the four components, with separate earlier-release fits for the two losses. For a signed profile $v\in\{-1,0,+1\}^4$, let $n(v)$ and $k_j(v)$ be its counts of earlier-release initial judgments and errors, and let $\bar p_j$ be the pooled earlier error rate. The profile risk estimate is $[k_j(v)+20\bar p_j]/[n(v)+20]$, retaining A/B orientation. Logistic risk uses 17 features: an intercept, four signed components, four absolute components, six pairwise products, the initial direction, and the absolute scalar margin. It minimizes summed binomial negative log-likelihood plus $\tfrac{1}{2}\|\beta_{\mathrm{nonintercept}}\|_2^2$, using float64 Newton iterations with maximum absolute gradient tolerance $10^{-7}$ and a limit of 100 iterations. Target ties remain in the directional fit with error indicator zero. No ranker uses demographic attributes; later target labels are used only for evaluation.

\begin{table}[!htbp]
\caption{Complete fixed-budget review comparison. Entries are expected error-capture percentages on the same 26,710 initial E.25 judgments. Budgets correspond to 267, 534, 1,335, 2,222, 2,671, 4,006, and 5,342 slots. Panel a uses 487 directional errors; panel b uses 2,035 tie-aware errors. Learned rankers use separate earlier-release fits for each panel's loss. All other priorities use component evidence alone. ``Pareto then margin'' sorts within both Pareto layers and is distinct from the equivalent release rule in \Cref{eq:humaine-hybrid}.}
\label{tab:review-budgets}
\centering\footnotesize
\setlength{\tabcolsep}{3pt}
\begin{tabular*}{\textwidth}{@{\extracolsep{\fill}}lrrrrrrr@{}}
\toprule
Review priority & 1\% & 2\% & 5\% & 8.32\% & 10\% & 15\% & 20\%\\
\midrule
\multicolumn{8}{@{}l}{\emph{a\quad Directional errors}}\\
Smallest margin & 3.21 & 6.41 & 16.03 & 26.68 & 32.07 & 48.10 & 64.14\\
Pareto frontier & 5.36 & 10.71 & 26.79 & 44.58 & 53.59 & 61.50 & 63.76\\
Partial frontier & 6.49 & 12.98 & 32.45 & 54.00 & 54.85 & 57.36 & 59.86\\
Pareto then margin & 6.49 & 12.98 & 32.45 & 54.00 & 57.42 & 66.01 & 74.15\\
Learned profile risk & 13.70 & 25.76 & 38.46 & 53.21 & 59.32 & 69.67 & 75.46\\
Learned logistic risk & 13.12 & 27.13 & 42.87 & 53.80 & 59.77 & 70.93 & 76.33\\
\addlinespace
\multicolumn{8}{@{}l}{\emph{b\quad Tie-aware errors}}\\
Smallest margin & 2.78 & 5.57 & 13.92 & 23.18 & 27.86 & 41.78 & 55.72\\
Pareto frontier & 2.40 & 4.79 & 11.98 & 19.95 & 23.98 & 29.91 & 34.03\\
Partial frontier & 2.85 & 5.70 & 14.26 & 23.73 & 25.13 & 29.29 & 33.45\\
Pareto then margin & 2.85 & 5.70 & 14.26 & 23.73 & 25.51 & 37.33 & 51.10\\
Learned profile risk & 4.50 & 8.45 & 20.41 & 32.21 & 36.99 & 50.68 & 62.71\\
Learned logistic risk & 3.79 & 7.84 & 20.22 & 32.46 & 37.30 & 50.11 & 61.30\\
\bottomrule
\end{tabular*}
\end{table}

\Cref{tab:review-budgets} reports complete point estimates across all budgets, including Pareto then margin; \Cref{tab:review-contrasts} gives reference-budget paired differences for the original binary priority. Five baselines, seven budgets, and two losses form a 70-comparison interval family; the exactly equivalent hybrid is excluded. These review comparisons are separate from the seven-candidate release-calibration family.

\begin{table}[!htbp]
\caption{Paired error-capture differences at the 2,222-slot reference budget. Values are percentage points, partial-frontier priority minus the named baseline. Intervals use 20,000 paired, group-stratified comparison bootstraps with budget reallocation. The last column adjusts percentile endpoints for 70 comparisons over seven budgets and two losses. All intervals condition on the earlier fits, group sizes, and budget.}
\label{tab:review-contrasts}
\centering\footnotesize
\setlength{\tabcolsep}{4pt}
\begin{tabular*}{\textwidth}{@{\extracolsep{\fill}}lrrr@{}}
\toprule
Baseline & Difference & Pointwise 95\% CI & Adjusted CI\\
\midrule
\multicolumn{4}{@{}l}{\emph{Directional errors}}\\
Uniform random & $+45.69$ & $[40.97,49.58]$ & $[37.30,52.56]$\\
Smallest margin & $+27.33$ & $[23.03,30.69]$ & $[20.05,33.42]$\\
Pareto frontier & $+9.42$ & $[6.68,11.20]$ & $[4.69,12.52]$\\
Learned profile risk & $+0.80$ & $[-3.40,4.18]$ & $[-6.18,7.00]$\\
Learned logistic risk & $+0.21$ & $[-1.21,0.88]$ & $[-2.35,1.68]$\\
\addlinespace
\multicolumn{4}{@{}l}{\emph{Tie-aware errors}}\\
Uniform random & $+15.42$ & $[13.58,17.00]$ & $[12.29,18.21]$\\
Smallest margin & $+0.56$ & $[-1.17,2.00]$ & $[-2.40,3.13]$\\
Pareto frontier & $+3.79$ & $[2.86,4.42]$ & $[2.29,4.89]$\\
Learned profile risk & $-8.47$ & $[-10.61,-6.57]$ & $[-12.07,-5.22]$\\
Learned logistic risk & $-8.73$ & $[-10.67,-7.09]$ & $[-12.22,-5.97]$\\
\bottomrule
\end{tabular*}
\end{table}

\subsection{Does within-layer ordering add information beyond the frontier?}
\label[appendix]{app:review-refinements}
To distinguish a structural set from a complete risk ranking, we evaluate six further priorities on the unchanged workload and budget grid (\Cref{tab:review-refinements}). Robust gap sorts by increasing $\underline u_{\Wp}(sx)$, where $s=\sign(w_0^\top x)$ is the initial direction. Partial then margin sorts within both partial layers by increasing scalar margin. Four learned variants sort within the partial or Pareto layers by decreasing values of the unchanged profile or logistic risk scores. These variants add no fitted parameters. Only the learned variants use earlier target labels. All final ties retain uniform inclusion probabilities.

By \Cref{eq:humaine-closed-form}, the robust gap is zero for every eligible partial-frontier profile. The next tier combines oriented $(1,0,0,0)$ and $(1,1,1,-1)$ profiles, both with gap .20. Pareto then margin orders the latter first; partial then margin orders the former first. Robust gap therefore retains more ordering information than a binary core indicator, but does not distinguish cases within the partial frontier.

\begin{table}[!htbp]
\caption{Review-ranking extensions on the same seven budgets as \Cref{tab:review-budgets}. Entries are expected capture percentages. Robust gap and partial then margin use no target-risk fit; the other variants use the unchanged earlier fits for each loss. The corresponding unrefined baselines and Pareto then margin are in \Cref{tab:review-budgets}.}
\label{tab:review-refinements}
\centering\footnotesize
\setlength{\tabcolsep}{3pt}
\begin{tabular*}{\textwidth}{@{\extracolsep{\fill}}lrrrrrrr@{}}
\toprule
Review priority & 1\% & 2\% & 5\% & 8.32\% & 10\% & 15\% & 20\%\\
\midrule
\multicolumn{8}{@{}l}{\emph{Directional errors}}\\
Robust gap & 6.49 & 12.98 & 32.45 & 54.00 & 56.83 & 65.25 & 73.67\\
Partial then margin & 6.49 & 12.98 & 32.45 & 54.00 & 56.74 & 64.87 & 73.01\\
Partial then profile & 13.70 & 25.76 & 39.43 & 54.00 & 60.08 & 71.59 & 77.64\\
Partial then logistic & 13.12 & 27.13 & 42.87 & 54.00 & 59.71 & 70.93 & 76.33\\
Pareto then profile & 13.70 & 25.76 & 39.43 & 52.09 & 54.96 & 72.84 & 79.55\\
Pareto then logistic & 13.12 & 27.13 & 42.87 & 53.72 & 56.29 & 72.72 & 79.47\\
\addlinespace
\multicolumn{8}{@{}l}{\emph{Tie-aware errors}}\\
Robust gap & 2.85 & 5.70 & 14.26 & 23.73 & 27.96 & 40.53 & 53.11\\
Partial then margin & 2.85 & 5.70 & 14.26 & 23.73 & 28.36 & 42.13 & 55.90\\
Partial then profile & 4.46 & 8.40 & 17.50 & 23.73 & 30.50 & 47.27 & 60.50\\
Partial then logistic & 3.44 & 7.40 & 18.44 & 23.73 & 30.85 & 47.27 & 58.82\\
Pareto then profile & 4.46 & 8.40 & 17.50 & 23.68 & 25.65 & 42.22 & 56.44\\
Pareto then logistic & 3.44 & 7.40 & 18.44 & 23.73 & 25.26 & 42.22 & 56.44\\
\bottomrule
\end{tabular*}
\end{table}

At 20\%, robust gap improves directional capture over the binary partial-frontier priority by 13.81 points, but differs from Pareto then margin by $-0.48$ points. \Cref{tab:review-refinement-contrasts} gives the paired uncertainty and both losses. The refinements do not establish a unique directional-ranking advantage of robust gap over within-layer margin ordering. For tie-aware errors, robust gap exceeds Pareto then margin but falls below partial then margin and both learned rankers. Learned cascades can also improve the binary priority. Their directional gains over the corresponding unlayered ranker do not establish consistent superiority over the matched Pareto cascade.

\begin{table}[!htbp]
\caption{Robust gap minus alternative priorities at 20\% review budget. Differences and adjusted intervals are in percentage points. Intervals use 50,000 paired group-stratified replicates and Bonferroni percentile endpoints over the full 182-comparison extension family. Intervals spanning zero do not establish equivalence.}
\label{tab:review-refinement-contrasts}
\centering\footnotesize
\setlength{\tabcolsep}{4pt}
\begin{tabular*}{\textwidth}{@{\extracolsep{\fill}}lrrrr@{}}
\toprule
& \multicolumn{2}{c}{Directional errors} & \multicolumn{2}{c}{Tie-aware errors}\\
\cmidrule(lr){2-3}\cmidrule(lr){4-5}
Baseline & Difference & Adjusted CI & Difference & Adjusted CI\\
\midrule
Partial frontier & $+13.81$ & $[10.08,17.82]$ & $+19.66$ & $[17.29,22.27]$\\
Pareto frontier & $+9.91$ & $[5.57,14.37]$ & $+19.08$ & $[16.54,21.75]$\\
Smallest margin & $+9.53$ & $[5.90,13.25]$ & $-2.60$ & $[-4.06,-0.97]$\\
Partial then margin & $+0.66$ & $[-1.48,3.27]$ & $-2.78$ & $[-3.72,-1.81]$\\
Pareto then margin & $-0.48$ & $[-2.37,1.08]$ & $+2.01$ & $[1.30,2.73]$\\
Learned profile risk & $-1.79$ & $[-6.84,3.45]$ & $-9.59$ & $[-12.26,-6.56]$\\
Learned logistic risk & $-2.66$ & $[-7.04,1.88]$ & $-8.19$ & $[-10.57,-5.56]$\\
\bottomrule
\end{tabular*}
\end{table}

\FloatBarrier

\section{Statistical Procedures and Scope of Inference}
\label[appendix]{sec:artifact}

\paragraph{TradeoffBench.}
Point estimates weight eligible source pairs equally. Ordinary-interface breadth uncertainty uses 2,000 percentile-bootstrap samples of latent tasks; controlled rates and contrasts use 20,000. Endpoint, factorial, and disjoint-label evidence results share the exchange audit's original-orientation Vector/disjoint observations. The main endpoint counts and intervals use all 1,345 mixed-sign pairs; the 849-pair diagnostics resample that subset separately. Exchange intervals use 20,000 task-cluster bootstrap replicates, keeping both orientations together. Their consistency denominator includes every pair in the stated population. These intervals describe latent-task sampling, not variation across model families or uniform sampling of distinct categorical vectors.

The majority and fixed-coordinate audit reuses these saved outputs without new inference. Its pair- and profile-weighted summaries (\Cref{app:endpoint-swap}) describe the archive; the 577 outcomes are model--pair events, not independent pairs. Exact rule agreement is separate from the task-bootstrap intervals for consistency frequency.

The exhaustive-profile summaries in \Cref{app:lattice-audit} concern repeated presentations of a finite grid, not independent draws of new profiles.
The ON mixed-sign denominator is $180\times2\times3\times3\times3=9,720$; the control denominator is $63\times2\times3\times3\times3=3,402$, covering orientations, encodings, seeds, and models.
Completion-length maxima are descriptive statistics of these runs.
The deterministic interface in \Cref{app:reasoning-integration} is a rule implementation, not a further empirical ON/OFF comparison.

\paragraph{MultiPref component substitution.}
The original own/peer-mean contrasts use 20,000 bootstrap replicates with crossed evaluator and prompt clusters. Both conditions receive the same weights, retaining within-record pairing. The shared-core contrast conditions on records released in both conditions, so its risk difference is not driven by core membership changes. The single-peer ablation uses 5,000 paired crossed-bootstrap replicates and Bonferroni-adjusted percentile intervals across its three risk contrasts (\Cref{app:single-peer}).

\paragraph{HUMAINE release and calibration.}
Earlier-release comparison counts yield exact one-sided Clopper--Pearson upper bounds, corrected over seven distinct candidates (42 group--candidate bounds for the shared worst-group contract). The later release supplies evaluation metrics. Coverage contrasts use 20,000 paired, group-stratified bootstraps conditional on the selected policies. For the 2.485-point partial-minus-Pareto coverage difference, the pointwise 95\% interval is $[2.311,2.662]$ and the simultaneous interval is $[2.084,2.886]$. The latter uses the configured maximum-deviation band across contract--loss--tolerance contrasts. Bonferroni and Holm sensitivities use the same finite families and \Cref{eq:calibration-test}. Effective counts are rounded before testing, and each assumed fraction is analyzed separately. Qualification assumes independent comparison units and release-to-release transport of the relevant population. Participant identifiers are unavailable: the effective-sample analysis is a hypothetical stress test, not an estimated participant-cluster correction.

\paragraph{HUMAINE review allocation.}
For \Cref{app:review-allocation}, 20,000 replicates resample initial strict judgments within the six country--age groups, preserving their observed sizes. Every original priority, budget, and loss shares the draws. Each replicate reallocates the fixed integer budget using unchanged scores and recomputes expected error capture. Five original baseline contrasts across seven budgets and two losses form the 70-comparison interval family. Paired intervals condition on the earlier-release fits, group sizes, and budgets; they do not include refitting on new earlier data.

\paragraph{Review-ranking extensions.}
The extensions use 50,000 shared replicates. Their 182-comparison family contains thirteen contrasts at seven budgets for two losses. Seven compare robust gap with the original partial, Pareto, margin, two learned, and two within-layer margin priorities. The other six compare each partial learned cascade with the original partial priority, its unlayered learned ranker, and its matched Pareto cascade. Each priority is recomputed separately. Partial frontier and Pareto then margin have identical inclusion probabilities by construction when the budget does not exceed the replicate's frontier capacity. Beyond that capacity, the priorities can assign different inclusion probabilities. Their equality at the observed 2,222-slot capacity therefore does not imply equality in every resample. Pointwise intervals use the 2.5th and 97.5th percentiles. For family size $M$, adjusted percentile probabilities are $.05/(2M)$ and $1-.05/(2M)$. These are comparison-level bootstrap intervals without a participant-cluster correction. Capture quantifies errors placed in a queue, not observed review outcomes.

\clearpage
\section{Response-Text Audits and Additional Validation}
\label[appendix]{app:revision-evidence}

\subsection{Reasoning-enabled judges on raw and open-ended responses}
\label[appendix]{app:raw-open}
The raw TradeoffBench audit evaluates 400 pairs in both orders: 200 verified conflicts and 100 each of dominance and equality controls, using the original raw-evidence/disjoint-label prompt. Qwen3-8B gives 84/400 unsupported strict verdicts and answers 196/200 dominance and 200/200 equality controls correctly. Qwen3.8-27B gives 31/400 unsupported strict verdicts and answers all 400 controls correctly. All 1,600 outputs are valid. This condition tests assessment and aggregation jointly, without authoritative component vectors.

\paragraph{Frontier judging with specified and unspecified trade-offs.}
GPT-6.1-sol evaluates the same 800 presentations through the Responses API, at medium reasoning effort, Standard service, and a 2,048-token cap including reasoning, without tools. The explicit-rule condition requires the Pareto verdict, forbids additional weights, priorities, and majority voting, and returns only a verdict tag. The unspecified condition requests the five component comparisons and an overall assessment, allows an explanation, and permits A, B, EQUAL, or UNDERDETERMINED. It uses the same questions, rubrics, response texts, presentation order, and inference parameters. The decision instruction and output format change together. Each protocol is fixed before generation; each presentation is generated once. All 1,600 outputs are complete and valid, with no retries.

Under the explicit rule, all 400 conflicts receive UNDERDETERMINED. With trade-offs unspecified, 176 receive a strict verdict, 220 receive UNDERDETERMINED, and four receive EQUAL. Both conditions answer all 400 controls correctly (\Cref{tab:raw-open}). In the unspecified condition, all 400 verified conflicts remain conflicts in the reported components, with no spurious conflicts on controls. Complete-vector matches number 351/400 on conflicts, 188/200 on dominance, and 200/200 on equality; format accounts for all component mismatches. Of the 176 strict conflict verdicts, 153 follow component reports matching the complete verified vector.

All 166 strict verdicts on non-tied verified profiles favor the criterion majority (\Cref{tab:frontier-profiles}). Across both orders, 73 conflict pairs receive the same strict candidate choice, 30 are strict in one order only, and 97 are strict in neither. All 73 consistently selected candidates are the verified majority. Equal weighting supplies a completion premise when the rubric leaves the trade-off unspecified (\Cref{sec:core-frontier}); a declared majority choice is an explicit completion.

\begin{table}[!htbp]
\caption{Strict verdicts by verified conflict profile for GPT-6.1-sol with unspecified trade-offs. A profile counts criteria favoring each candidate, ignoring equal criteria and display order.}
\label{tab:frontier-profiles}
\centering\footnotesize
\begin{tabular}{lrrr}
\toprule
Profile & Presentations & Strict verdicts & Rate (\%)\\
\midrule
1 vs.\ 1 & 148 & 8 & 5.4\\
2 vs.\ 2 & 24 & 2 & 8.3\\
2 vs.\ 1 & 156 & 103 & 66.0\\
3 vs.\ 2 & 6 & 5 & 83.3\\
3 vs.\ 1 & 42 & 35 & 83.3\\
4 vs.\ 1 & 24 & 23 & 95.8\\
\bottomrule
\end{tabular}
\end{table}

The disclosure criterion in \Cref{app:realtext-diagnostic} gives 52 explicit premises, 119 unstated premises, and five ambiguous cases. The explicit group contains 33 majority/counting premises, 12 criterion-specific reasons, and seven outweighing relations. The unstated group contains 52 outputs with no explanatory prose, 46 with criterion descriptions only, and 21 with a bare overall preference but no stated aggregation premise. Among the 52 with no prose, 49 match the full verified vector. Unstated-premise verdicts constitute 29.8\% of the 400 conflicts (95\% CI $[24.3,35.6]$); those with no prose constitute 13.0\% ($[9.6,16.8]$). Intervals use 10,000 bootstrap draws over the 173 latent task clusters containing conflicts, retaining both orders together. They describe task sampling with the recorded outputs and disclosure categories held fixed. These results concern one frontier judge on mechanically verifiable synthetic tasks; the open-ended audits below use MultiPref responses.

\paragraph{Open-ended component assessment.}

The open-ended audit uses 700 MultiPref pairs: 300 reference conflicts, 200 dominance cases, and 200 equalities, spanning 662 prompts and 184 target evaluators. Both orders are evaluated under helpfulness, truthfulness, and harmlessness, with human labels withheld. Judges report components and a verdict under the strict Pareto rule, using EQUAL for equal components and UNDERDETERMINED for mixed signs. Additional weights, priorities, majority voting, and severity overrides are forbidden. An aggregation violation is a verdict inconsistent with the rule applied to that output's own components; a human-reference conflict alone is insufficient.

Both Qwen models use BF16, vLLM, a 16,384-token context, and an 8,192-token output cap. Reasoning is enabled, with temperature .6 for Qwen3-8B and temperature 1 at medium effort for Qwen3.8-27B; both use top-$p$ .95 and top-$k$ 20. The exhaustive-vector audit uses a 2,048-token cap. The cross-audit contrast therefore does not isolate input format.

\begin{table}[!htbp]
\caption{Open-ended component audit on 700 pairs in both orders. Normalization adds only missing outer JSON braces to an otherwise complete component object. It changes no comparison or verdict. Unresolved outputs are retained in request-level denominators and are not counted as successful deferrals.}
\label{tab:open-parsing}
\centering\footnotesize
\setlength{\tabcolsep}{4pt}
\begin{tabular*}{\textwidth}{llrrrr@{}}
\toprule
Judge & Parsing & Valid / requests & Reported conflicts & Strict on conflicts & All rule mismatches\\
\midrule
Qwen3-8B & Normalized & 1,399 / 1,400 & 196 & 7 & 7\\
& Strict JSON & 181 / 1,400 & 24 & 1 & 1\\
Qwen3.8-27B & Normalized & 1,399 / 1,400 & 146 & 0 & 0\\
& Strict JSON & 1,335 / 1,400 & 143 & 0 & 0\\
\bottomrule
\end{tabular*}
\end{table}

Six of the seven Qwen3-8B violations require brace normalization; all seven occur on distinct pairs and prompts. One output per judge is unresolved. The initial reference is one recorded rater; \Cref{app:realtext-diagnostic} uses four-rater consensus. These deliberately stratified samples and paired orientations do not estimate population prevalence or constitute independent replicates.

\subsection{Open-ended judgments under unspecified trade-offs}
\label[appendix]{app:realtext-diagnostic}
We use all 73 pairs in the 700-pair audit whose entire helpfulness, truthfulness, and harmlessness sign vector agrees across four distinct recorded raters: 23 conflicts, 23 dominance cases, and 27 equalities, each from a distinct prompt. Selection uses human annotations, not model outcomes. Sign consensus need not imply equal ordinal magnitudes.

The model reads the original request and both complete responses. Four conditions cross component source with aggregation instructions. Components are either assessed with human labels withheld or copied from the fixed human reference without rescoring. Trade-offs are unspecified or governed by the explicit strict Pareto rule. The latter requires EQUAL for equal components and UNDERDETERMINED for mixed signs, forbidding additional weights, priorities, severity overrides, and majority voting. All conditions permit A, B, EQUAL, UNDERDETERMINED, and an explanation; component labels are A\_ONLY, SAME, and B\_ONLY.

GPT-6.1-sol is a hosted alias accessed through the Responses API at medium reasoning effort, Standard service, and a 2,048-token cap. Both orders give 146 requests per condition. Of 584 requests, 581 return valid, complete outputs; three service rejections on one pair remain unresolved.

\paragraph{Disclosure of a completion premise.}
The audit reads each complete final answer, including text before and after the component object. A stated priority, outweighing relation, or case-specific reason linking the overall choice to a criterion counts as an explicit premise; numerical weights are unnecessary. An overall preference despite an acknowledged criterion disadvantage also counts as explicit. Criterion descriptions or a bare overall preference without an express aggregation premise are coded as unstated, with ambiguous wording separate. An output containing only components and a verdict has no explanatory prose. The endpoint is disclosure in the returned final answer.

\begin{table}[!htbp]
\caption{GPT-6.1-sol on 73 MultiPref pairs with four-rater component consensus, each in both orders. Conflicts use reported components in assessed conditions and supplied components in fixed conditions. Explicit, unstated, and ambiguous classify the completion premise among strict conflict verdicts. Three of 584 requests are unresolved; all returned outputs are valid and complete.}
\label{tab:realtext-diagnostic}
\centering\footnotesize
\setlength{\tabcolsep}{3pt}
\begin{tabular*}{\textwidth}{@{\extracolsep{\fill}}llrrrrrr@{}}
\toprule
Components & Trade-offs & \shortstack{Valid /\\planned} & Conflicts & Strict & Explicit & Unstated & Ambiguous\\
\midrule
Assessed & Unspecified & 145 / 146 & 37 & 36 & 29 & 6 & 1\\
Assessed & Explicit rule & 145 / 146 & 38 & 0 & 0 & 0 & 0\\
Fixed & Unspecified & 145 / 146 & 45 & 36 & 36 & 0 & 0\\
Fixed & Explicit rule & 146 / 146 & 46 & 0 & 0 & 0 & 0\\
\bottomrule
\end{tabular*}
\end{table}

The six unstated-premise outputs span five prompts; two match the complete four-rater sign vector. All occur under unspecified trade-offs.

\paragraph{Separating aggregation from component assessment.}
Among 118 presentations with identical assessed vectors across the two instruction conditions, 31 are conflicts; strict verdicts change from 30 to zero. All valid fixed-component copies are correct, and both fixed-component conditions preserve all 100 control verdicts. Control-verdict agreement in the assessed-component conditions is 62/100 versus 50/100, reflecting both measurement and aggregation.

\paragraph{Human-reference conflicts under the explicit rule.}
\Cref{tab:consensus-measurement} \rev{in \Cref{sec:raw-open}} restricts model-assessed explicit-rule outputs to the 23 unanimous-sign conflict pairs. GPT-6.1-sol and Qwen3.5-9B use the four-condition audits; Qwen3-8B and Qwen3.8-27B use the corresponding pairs from the 700-pair audit (\Cref{app:raw-open}).

Every strict verdict follows dominance in the model's reported components, whereas the human reference has mixed signs. One GPT request is rejected by the service. Qwen3.5-9B has 44 actual generations: two planned presentations are ungenerated and 15 generated outputs are unresolved under primary parsing. Literal-field sensitivity resolves 39/44, comprising 34 strict verdicts and five conflicts, with zero rule mismatches.

Among the 23 unanimous-sign conflict pairs, seven have opposing strict overall labels; four retain opposing labels among raters with identical full ordinal components. No dominance/equality pair has opposing strict labels. This documents non-uniqueness in the recorded mapping.

\subsection{Open-weight replication and output-format sensitivity}
\label[appendix]{app:frontier}
Qwen3.5-9B repeats the four-condition audit using BF16 revision \texttt{\seqsplit{c202236235762e1c871ad0ccb60c8ee5ba337b9a}}, thinking enabled, and model-card general-task sampling: temperature 1, top-$p$ .95, top-$k$ 20, and presence penalty 1.5. Context and output limits are 16,384 and 8,192 tokens. One pair associated with a hosted-service safety rejection remains ungenerated locally without replacement: eight planned entries remain unresolved, leaving 72 pairs and 576 generations.

\begin{table}[!htbp]
\caption{Qwen3.5-9B replication. Primary parsing accepts complete component objects and verdict tags, with outer-brace normalization only. Literal-field sensitivity accepts uniquely stated criterion labels despite malformed wrappers; it does not infer labels from prose. All denominators are 144 actual generations. Strict conflict verdicts are unchanged by this sensitivity.}
\label{tab:frontier}
\centering\footnotesize
\begin{tabular*}{\textwidth}{@{\extracolsep{\fill}}llrrr@{}}
\toprule
Components & Trade-offs & Primary valid & Literal-field valid & Strict on conflict\\
\midrule
Assessed & Unspecified & 107 & 126 & 3\\
Assessed & Explicit rule & 100 & 128 & 0\\
Fixed & Unspecified & 138 & 144 & 44\\
Fixed & Explicit rule & 136 & 144 & 0\\
\bottomrule
\end{tabular*}
\end{table}

All 47 auditable strict conflict verdicts expressly state a priority or trade-off. All valid fixed-component copies are correct. The assessed conditions contain three and seven conflicts under primary parsing, or three and eight under literal-field sensitivity. The explicit-rule assessed arm also gives EQUAL on $(0,1,0)$, a non-conflict rule mismatch.

The literal-field parser accepts uniquely stated criterion labels and identical repeated fields, rejects contradictions, and maps two explicit SAME verdicts to EQUAL. It excludes capped outputs and does not infer labels from prose or reasoning traces. It recovers 61 outputs without changing primary valid labels or strict conflict counts. Thirty-four outputs remain unresolved, including one capped response, and are not counted as successful deferrals.

Under literal-field sensitivity, fixed-component control verdicts are correct in all 100 presentations under both instruction conditions. Thus aggregation changes on conflicts are isolated by the fixed comparison: all 44 supplied conflicts receive strict verdicts with stated priorities under unspecified trade-offs and defer under the explicit rule.

\subsection{Support checking with LLM-generated components}
\label[appendix]{app:workflow}
We compare a reasoning-enabled joint components-plus-verdict baseline with component-only assessment followed by deterministic identification. Balanced component agreement is mean recall over the nine criterion-by-reference-sign cells, counting unresolved outputs as incorrect. A fixed hash chooses one recorded rater per pair as the target.

Prompts are fixed using 80 development pairs per model; confirmation prompts are disjoint from development and earlier judge audits. Qwen3.8-27B uses the BF16 settings in \Cref{app:raw-open} on 200 pairs: 120 dominance, 48 conflict, and 32 equality cases. GPT-6.1-sol uses the settings in \Cref{app:realtext-diagnostic} on 120 pairs: 72 dominance, 29 conflict, and 19 equality cases. Both orders are evaluated. The GPT prompts share sign labels and criterion definitions, differing only in joint versus component-only assessment. Comparisons are within each model.

\begin{table}[!htbp]
\caption{Confirmation comparisons with LLM-generated components. Balanced and ordinary agreement are percentages. Each pair is evaluated in both orders. The last column gives releases on strict targets / directional errors; release membership may differ between conditions.}
\label{tab:workflow}
\centering\footnotesize
\setlength{\tabcolsep}{3pt}
\begin{tabular*}{\textwidth}{@{\extracolsep{\fill}}llrrrr@{}}
\toprule
Judge & Assessment & Valid / calls & Balanced & Ordinary & Strict rel./err.\\
\midrule
Qwen3.8-27B & Joint & 400 / 400 & 50.70 & 68.25 & 257 / 93\\
& Component-only & 396 / 400 & 50.50 & 67.33 & 253 / 94\\
GPT-6.1-sol & Joint & 240 / 240 & 51.65 & 63.75 & 153 / 55\\
& Component-only & 240 / 240 & 52.24 & 62.64 & 153 / 56\\
\bottomrule
\end{tabular*}
\end{table}

Paired crossed prompt/evaluator bootstraps use 5,000 draws. Component-only minus joint balanced agreement is $-0.19$ points for Qwen3.8-27B (95\% CI $[-4.40,3.73]$) and $+0.59$ for GPT-6.1-sol ($[-5.20,7.59]$). One GPT draw lacks a reference cell, leaving 4,999 estimable draws. GPT complete-vector matches are 64/240 versus 59/240; its 149 commonly released strict targets have 53 versus 55 errors. These confirmation results do not establish improved component or target agreement.

\paragraph{Enforcing support on unchanged components.}
On the 700-pair audit, deterministic identification changes seven Qwen3-8B strict conflict verdicts to deferral, preserving all other valid verdicts. Qwen3.8-27B remains unchanged. Of five changed presentations with strict targets, three native verdicts disagree with the target and two agree. Six repairs require brace normalization; strict-format parsing retains one. The guarantee concerns support under the declared rule for the reported components.

\subsection{Eliciting and validating an admissible weight set}
\label[appendix]{app:elicitation}
Reference comparisons constrain weights for intended evaluators, conditional on the declared criteria, scales, and additive model \citep{greco2008ordinal}. Noncompensatory requirements such as vetoes or lexicographic priorities need a different specification.

We use MultiPref's existing global-time and prompt-disjoint split and each evaluator's own ordinal components, coded as $\{-1,-.5,0,.5,1\}$. Queries are the earliest $k$ mixed-sign development comparisons, ordered by timestamp and comparison key without inspecting overall labels; all available mixed-sign comparisons are used if fewer than $k$ exist. Evaluation covers 4,847 later records from 80 evaluators.

Starting with $\mathbf1^\top w=1$ and $w_j\geq.01$, each queried strict preference $y_i\in\{-1,+1\}$ imposes
\begin{equation}
y_i w^\top x^{(i)}\geq10^{-6}.
\label{eq:elicited-constraint}
\end{equation}
Target ties consume a query without imposing a directional constraint or exact utility equality. All queried constraints are recorded. If their intersection is empty, we record the incompatibility and fall back to the initial set; otherwise, vertex enumeration supplies utility extrema, with release requiring worst-case signed utility above $10^{-8}$. On the observed profiles, the initial set yields the same decisions as strict Pareto, and refinement preserves every baseline release.

The primary four-query cap acquires 180 labels, 164 strict. The resulting sets are nonempty for 78/80 evaluators, including fourteen with no directional query information; two infeasible sets trigger the recorded fallback. The policy adds 110 decisions (+2.269 coverage points; descriptive 95\% CI $[0.992,3.819]$ from 5,000 crossed evaluator/prompt draws). The added layer contains 16 errors among 106 released strict targets, a strict-direction risk of 15.1\%, plus four target ties. Complete-policy risk is 39/3,375 (1.156\%), versus Pareto's 23/3,269 (0.704\%). Elicitation therefore expands coverage here at higher observed disagreement, especially on the added layer \rev{(\Cref{tab:elicitation} in \Cref{sec:elicitation})}.

Weight floors .001, .01, and .05 yield identical decisions at each query cap. Independent linear programming verifies primary-condition feasibility and utility extrema. The added decisions require separate target-risk validation before qualification. HUMAINE uses declared specifications.

\end{document}